\documentclass[10pt,twocolumn,letterpaper]{article}

\usepackage{iccv}
\usepackage{times}
\usepackage{epsfig}
\usepackage{graphicx}
\usepackage{amsmath}
\usepackage{amssymb}
\usepackage{amsthm}
\usepackage{subfig}
\usepackage{tabularx}
\usepackage{booktabs}
\usepackage{multirow}
\usepackage{multicol}
\usepackage{enumitem}
\usepackage{array}
\usepackage{color, colortbl}
\usepackage{fixltx2e}

\usepackage[pagebackref=true,breaklinks=true,letterpaper=true,colorlinks,bookmarks=false]{hyperref}

\definecolor{Gray}{gray}{0.92}
\definecolor{R}{rgb}{1.0,0.00,0.00}
\definecolor{Y}{gray}{0.85}
\definecolor{G}{gray}{1.0}

\newtheorem{property}{Property}
\newtheorem{note}{Note}
\newcommand{\bx}{\mathbf{x}}

\newcommand{\bz}{\mathbf{z}}
\newcommand{\bmu}{\boldsymbol{\mu}}
\newcommand{\bsigma}{\boldsymbol{\sigma}}
\newcommand{\hmu}{\hat{\mu}}
\newcommand{\hsigma}{\hat{\sigma}}

\newcommand{\cX}{\mathcal{X}}
\newcommand{\cZ}{\mathcal{Z}}
\newcommand{\norm}[1]{\left\lVert#1\right\rVert}

\newcommand{\IE}{\mathbb{E}}
\newcommand{\EL}{\mathcal{L}}

\iccvfinalcopy % *** Uncomment this line for the final submission

\def\iccvPaperID{6328} % *** Enter the ICCV Paper ID here

\ificcvfinal\fi
\begin{document}

%%%%%%%%% TITLE
\title{Probabilistic Face Embeddings}

% \author{
%     Yichun Shi\IEEEauthorrefmark{1},\;
%     Anil K. Jain\IEEEauthorrefmark{1},\;
%     Nathan D. Kalka\IEEEauthorrefmark{2}\\
%     \IEEEauthorrefmark{1}Michigan State University, East Lansing, MI\\
%     \IEEEauthorrefmark{2}Noblis, Bridgeport, WV\\
% {\tt\small shiyichu@msu.edu, jain@cse.msu.edu, nathan.kalka@noblis.org }
% }
\author{
    Yichun Shi\; and\; Anil K. Jain\\
    Michigan State University, East Lansing, MI\\
{\tt\small shiyichu@msu.edu, jain@cse.msu.edu}
}

\maketitle
%\thispagestyle{empty}

%%%%%%%%% ABSTRACT
\begin{abstract}
   Embedding methods have achieved success in face recognition by comparing facial features in a latent semantic space. However, in a fully unconstrained face setting, the facial features learned by the embedding model could be ambiguous or may not even be present in the input face, leading to noisy representations. We propose \textit{Probabilistic Face Embeddings (PFEs)}, which represent each face image as a Gaussian distribution in the latent space. The mean of the distribution estimates the most likely feature values while the variance shows the uncertainty in the feature values. Probabilistic solutions can then be naturally derived for matching and fusing PFEs using the uncertainty information. Empirical evaluation on different baseline models, training datasets and benchmarks show that the proposed method can improve the face recognition performance of deterministic embeddings by converting them into PFEs. The uncertainties estimated by PFEs also serve as good indicators of the potential matching accuracy, which are important for a risk-controlled recognition system.
\end{abstract}

%%%%%%%%% BODY TEXT
\vspace{-1.0em}\section{Introduction}

% Most existing face recognition systems are embedding-based, which maps input images to a semantic embedding space. The face embedding is compact, disentangled and discriminative such that each embedding location should uniquely represent the identity-salient face features in the input images but do not encode any identity-irrelevant information. 
% Thus, a distance metric can be applied to measure the likelihood of a pair of images belonging to the same person. With recent success of deep learning, the embedding function is usually a deep CNN trained on large-scale datasets~\cite{taigman2014deepface,schroff2015facenet,wen2016discriminative,liu2017sphereface}.

When humans are asked to describe a face image, they not only give the description of the facial attributes, but also the confidence associated with them. For example, if the eyes are blurred in the image, a person will keep the eye size as an uncertain information and focus on other features. Furthermore, if the image is completely corrupted and no attributes can be discerned, the subject may respond that he/her cannot identify this face. This kind of uncertainty (or confidence) estimation is common and important in human decision making.

\begin{figure}[t]
    \centering
    \captionsetup{font=footnotesize}
    % \subfloat[example failures of embedding representation]{\includegraphics[width=0.49\linewidth]{fig/question_1.pdf}\quad
    %     \includegraphics[width=0.49\linewidth]{fig/question_2.pdf}}\\
    \subfloat[deterministic embedding]{\label{fig:front_a}\includegraphics[width=0.48\linewidth]{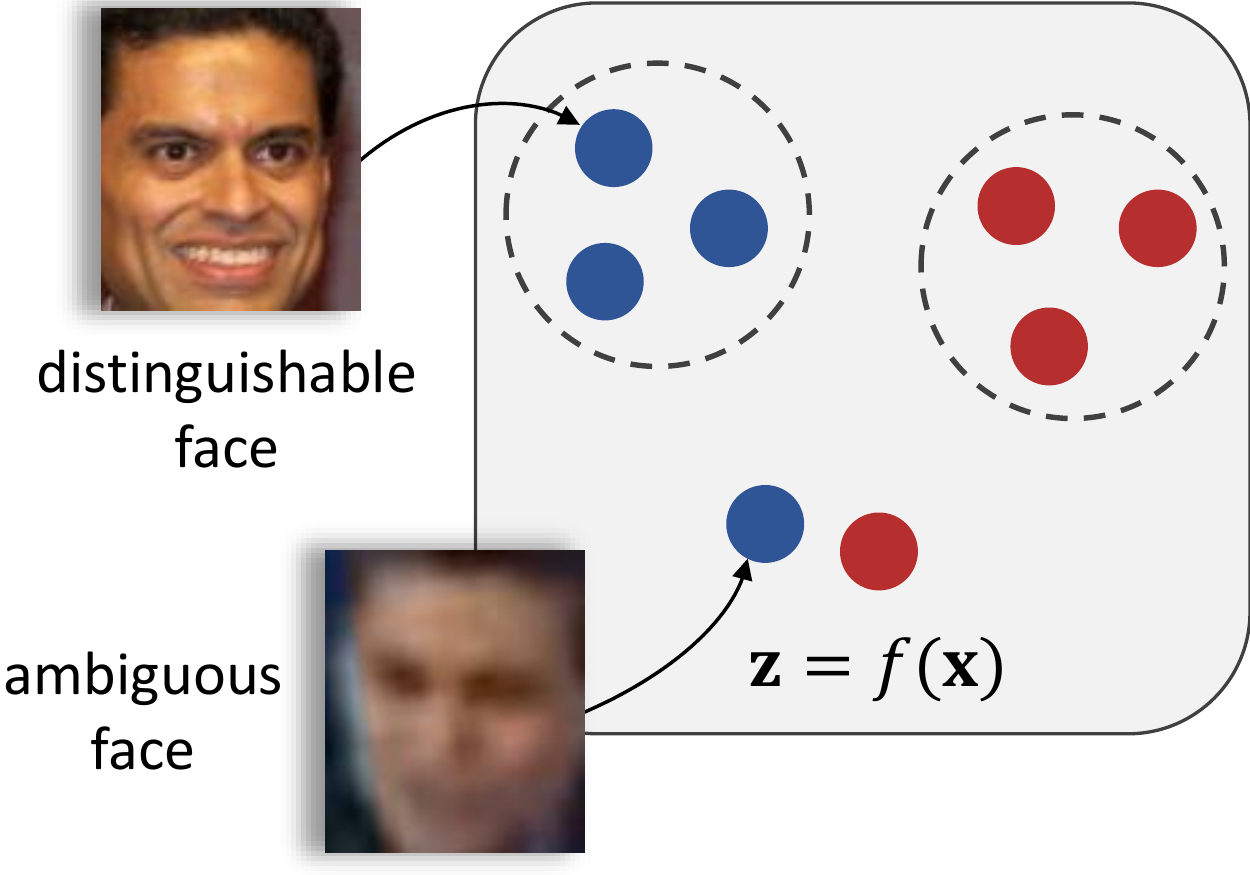}}\hfill
\subfloat[probabilistic embedding]{\label{fig:front_b}\includegraphics[width=0.48\linewidth]{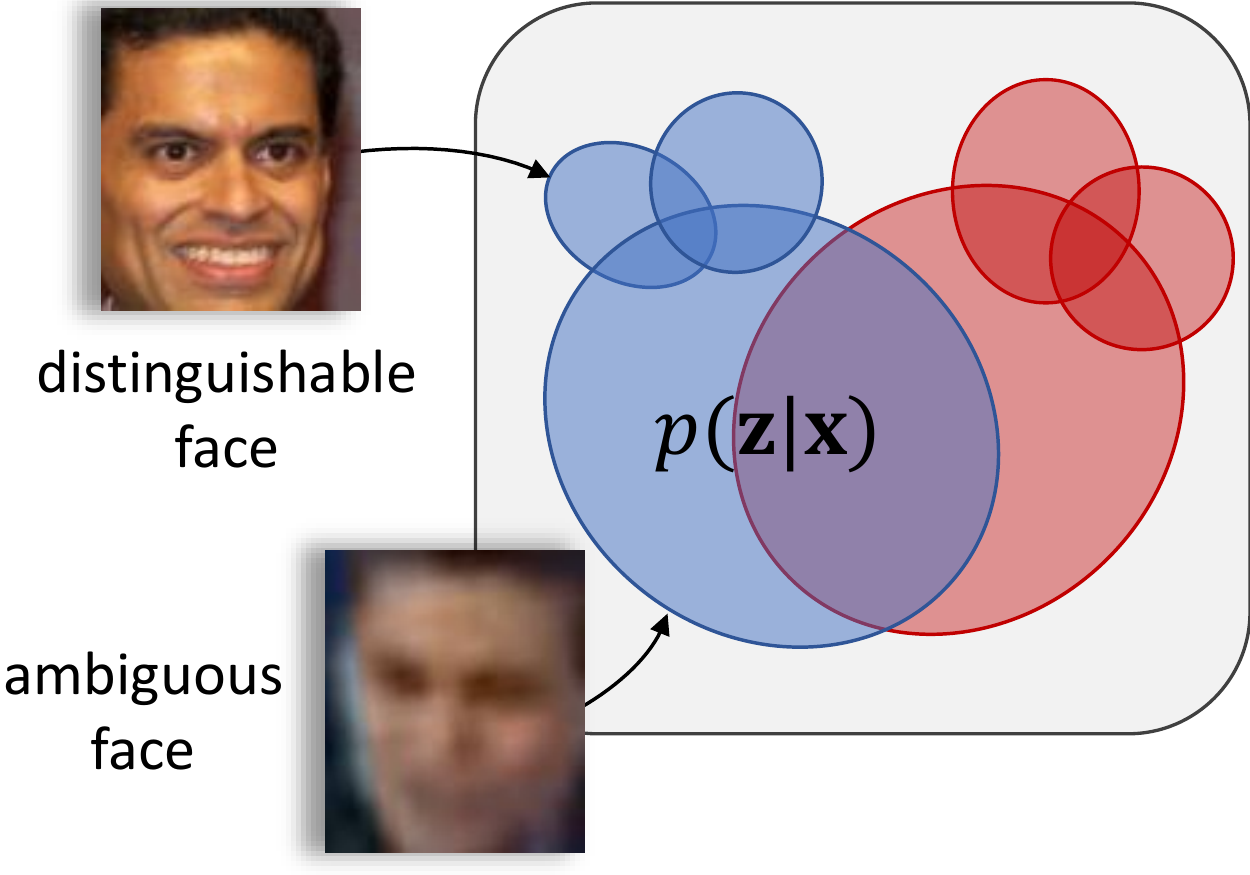}}
    \caption{Difference between deterministic face embeddings and probabilistic face embeddings (PFEs). Deterministic embeddings represent every face as a point in the latent space without regards to its feature ambiguity. Probabilistic face embedding (PFE) gives a distributional estimation of features in the latent space instead. \textbf{Best viewed
in color.}}\vspace{-1.5em}
    \label{fig:frontpage}
\end{figure}

On the other hand, the representations used in state-of-the-art face recognition systems are generally confidence-agnostic. These methods depend on an embedding model (\eg Deep Neural Networks) to give a deterministic point representation for each face image in the latent feature space~\cite{schroff2015facenet,wen2016discriminative,liu2017sphereface,wang2018cosface,deng2018arcface}. A point in the latent space represents the model's estimation of the facial features in the given image. If the error in the estimation is somehow bounded, the distance between two points can effectively measure the semantic similarity between the corresponding face images. But given a low-quality input, where the expected facial features are ambiguous or absent in the image, a large shift in the embedded points is inevitable, leading to false recognition (Figure~\ref{fig:front_a}).

\begin{figure}
    \centering
    \footnotesize
    \captionsetup{font=footnotesize}
    \newcommand{\vshrink}{\vspace{-10px}}
    \begin{minipage}{0.48\linewidth}
    \includegraphics[width=0.25\linewidth]{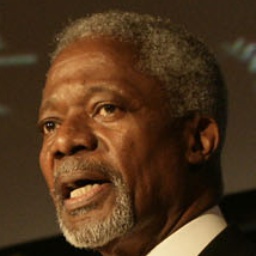}\hfill
    \includegraphics[width=0.25\linewidth]{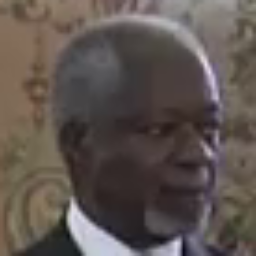}\hfill
    \includegraphics[width=0.25\linewidth]{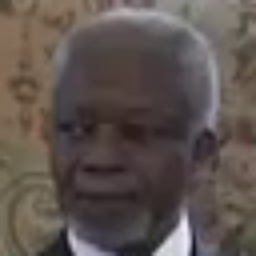}\hfill
    \includegraphics[width=0.25\linewidth]{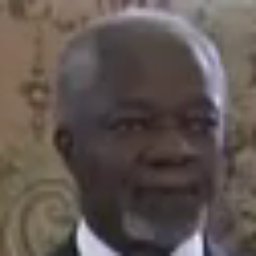}\\
    \includegraphics[width=0.25\linewidth]{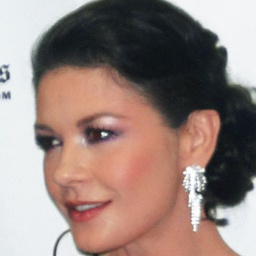}\hfill
    \includegraphics[width=0.25\linewidth]{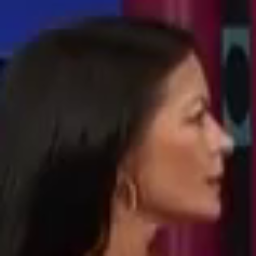}\hfill
    \includegraphics[width=0.25\linewidth]{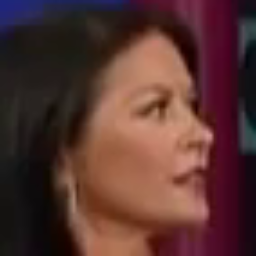}\hfill
    \includegraphics[width=0.25\linewidth]{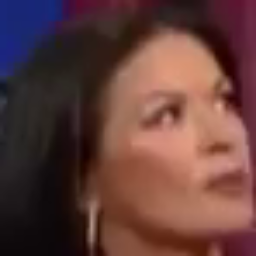}
    \vspace{-2.0em}\begin{center}(a) IJB-A~\cite{IJBA}\end{center}
    \end{minipage}\hfill
    \begin{minipage}{0.48\linewidth}
    \includegraphics[width=0.25\linewidth]{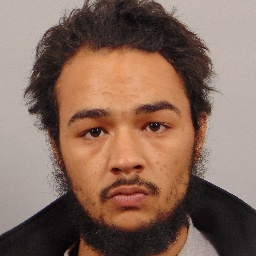}\hfill
    \includegraphics[width=0.25\linewidth]{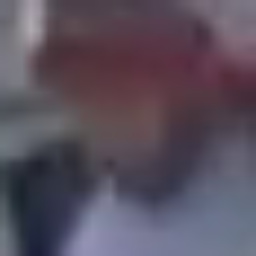}\hfill
    \includegraphics[width=0.25\linewidth]{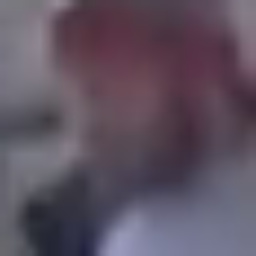}\hfill
    \includegraphics[width=0.25\linewidth]{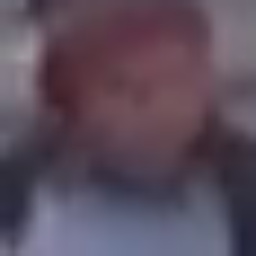}\\
    \includegraphics[width=0.25\linewidth]{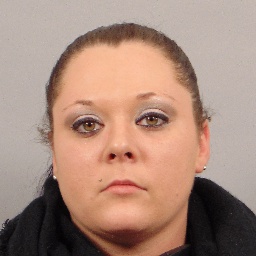}\hfill
    \includegraphics[width=0.25\linewidth]{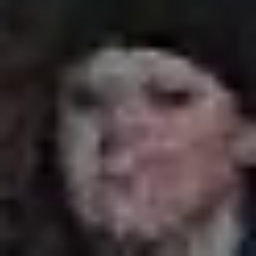}\hfill
    \includegraphics[width=0.25\linewidth]{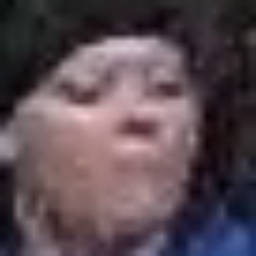}\hfill
    \includegraphics[width=0.25\linewidth]{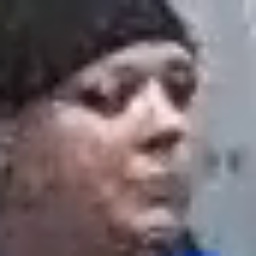}
    \vspace{-2.0em}\begin{center}(b) IJB-S~\cite{IJBS}\end{center}
    \end{minipage}\\
    \vspace{-0.9em}\caption{Example images from IJB-A and IJB-S. The first columns show still images, followed by video frames of the respective subjects in the next three columns. These benchmarks present a more unconstrained recognition scenario where there is a large variability in the image quality.}\vspace{-1.6em}
    \label{fig:dataset}
\end{figure}

Given that face recognition systems have already achieved high recognition accuracies on relatively constrained face recognition benchmarks,~\eg LFW~\cite{LFWTech} and YTF~\cite{YTF}, where most facial attributes can be clearly observed, recent face recognition challenges have moved on to more unconstrained scenarios, including surveillance videos~\cite{IJBA,IJBC,IJBS} (See Figure~\ref{fig:dataset}). In these tasks, any type and degree of variation could exist in the face image, where most of the desired facial features learned by the representation model could be absent. Given this lack of information, it is unlikely to find a feature set that could always match these faces accurately. Hence state-of-the-art face recognition systems which obtained over $99\%$ accuracy on LFW have suffered from a large performance drop on IARPA Janus benchmarks~\cite{IJBA,IJBC,IJBS}.

To address the above problems, we propose \textit{Probabilistic Face Embeddings (PFEs)}, which give a distributional estimation instead of a point estimation in the latent space for each input face image (Figure~\ref{fig:front_b}). The mean of the distribution can be interpreted as the most likely latent feature values while the span of the distribution represents the uncertainty of these estimations. PFE can address the unconstrained face recognition problem in a two-fold way: (1) During matching (face comparison), PFE penalizes uncertain features (dimensions) and pays more attention to more confident features. (2) For low quality inputs, the confidence estimated by PFE can be used to reject the input or actively ask for human assistance to avoid false recognition. Besides, a natural solution can be derived to aggregate the PFE representations of a set of face images into a new distribution with lower uncertainty to increase the recognition performance. The implementation of PFE is open-sourced\footnote{\url{https://github.com/seasonSH/Probabilistic-Face-Embeddings}}. The contributions of the paper can be summarized as below:
\begin{enumerate}[leftmargin=*]\vspace{-0.5em}
    \item An uncertainty-aware probabilistic face embedding (PFE) which represents face images as distributions instead of points.\vspace{-0.5em}
    \item A probabilistic framework that can be naturally derived for face matching and feature fusion using PFE.\vspace{-0.5em}
    \item A simple method that converts existing deterministic embeddings into PFEs without additional training data.\vspace{-0.5em}
    \item Comprehensive experiments showing that the proposed PFE can improve face recognition performance of deterministic embeddings and can effectively filter out low-quality inputs to enhance the robustness of face recognition systems.\vspace{-0.5em}
\end{enumerate}
% even some unconstrained benchmarks~\cite{IJBA}~\cite{IJBC}. 

\section{Related Work}

\paragraph{Uncertainty Learning in DNNs}
To improve the robustness and interpretability of discriminant Deep Neural Networks (DNNs), deep uncertainty learning is getting more attention~\cite{kendall2015bayesian,gal2016dropout,kendall2017uncertainties}. There are two main types of uncertainty: \textit{model uncertainty} and \textit{data uncertainty}. Model uncertainty refers to the uncertainty of model parameters given the training data and can be reduced by collecting additional training data~\cite{mackay1992practical,neal1995bayesian,kendall2015bayesian,gal2016dropout}. Data uncertainty accounts for the uncertainty in output whose primary source is the inherent noise in input data and hence cannot be eliminated with more training data~\cite{kendall2017uncertainties}. The uncertainty studied in our work can be categorized as data uncertainty. Although techniques have been developed for estimating data uncertainty in different tasks, including classification and regression~\cite{kendall2017uncertainties}, they are not suitable for our task since our target space is not well-defined by given labels\footnote{Although we are given the identity labels, they cannot directly serve as target vectors in the latent feature space.}. Variational Autoencoders~\cite{kingma2013auto} can also be regarded as a method for estimating data uncertainty, but it mainly serves a generation purpose. Specific to face recognition, some studies~\cite{gong2017capacity,khan2019striking,zafar2019face} have leveraged the model uncertainty for analysis and learning of face representations, but to our knowledge, ours is the first work that utilizes data uncertainty\footnote{Some in the literature have also used the terminology ``data uncertainty" for a different purpose~\cite{xu2014data}.} for recognition tasks.

% However, the output space of these tasks are well-defined by the given labels. To the best of our knowledge, this is the first work on data uncertainty in deep face representations, where the output space is a latent feature space that also needs to be modeled.

\vspace{-0.8em}\paragraph{Probabilistic Face Representation}
Modeling faces as probabilistic distributions is not a new idea. In the field of face template/video matching, there exists abundant literature on modeling the faces as probabilistic distributions~\cite{shakhnarovich2002face,arandjelovic2005face}, subspace~\cite{cevikalp2010face} or manifolds~\cite{arandjelovic2005face,huang2015log} in the feature space. However, the input for such methods is a set of face images rather than a single face image, and they use a between-distribution similarity or distance measure,~\eg KL-divergence, for comparison, which does not penalize the uncertainty. Meanwhile, some studies~\cite{li2013probabilistic,hiremath2007modelling} have attempted to build a fuzzy model of a given face using the features of face parts. In comparison, the proposed PFE represents each single face image as a distribution in the latent space encoded by DNNs and we use an uncertainty-aware log likelihood score to compare the distributions.

\vspace{-0.8em}\paragraph{Quality-aware Pooling}
In contrast to the methods above, recent work on face template/video matching aims to leverage the saliency of deep CNN embeddings by aggregating the deep features of all faces into a single compact vector~\cite{yang2017neural,liu2017quality,xie2018multicolumn,gong2019video}. In these methods, a separate module learns to predict the quality of each face in the image set, which is then normalized for a weighted pooling of feature vectors. We show that a solution can be naturally derived under our framework, which not only gives a probabilistic explanation for quality-aware pooling methods, but also leads to a more general solution where an image set can also be modeled as a PFE representation.

\begin{figure*}[t]
    \centering
    \captionsetup{font=footnotesize}
    \begin{minipage}{0.33\linewidth}
    \includegraphics[width=1\linewidth]{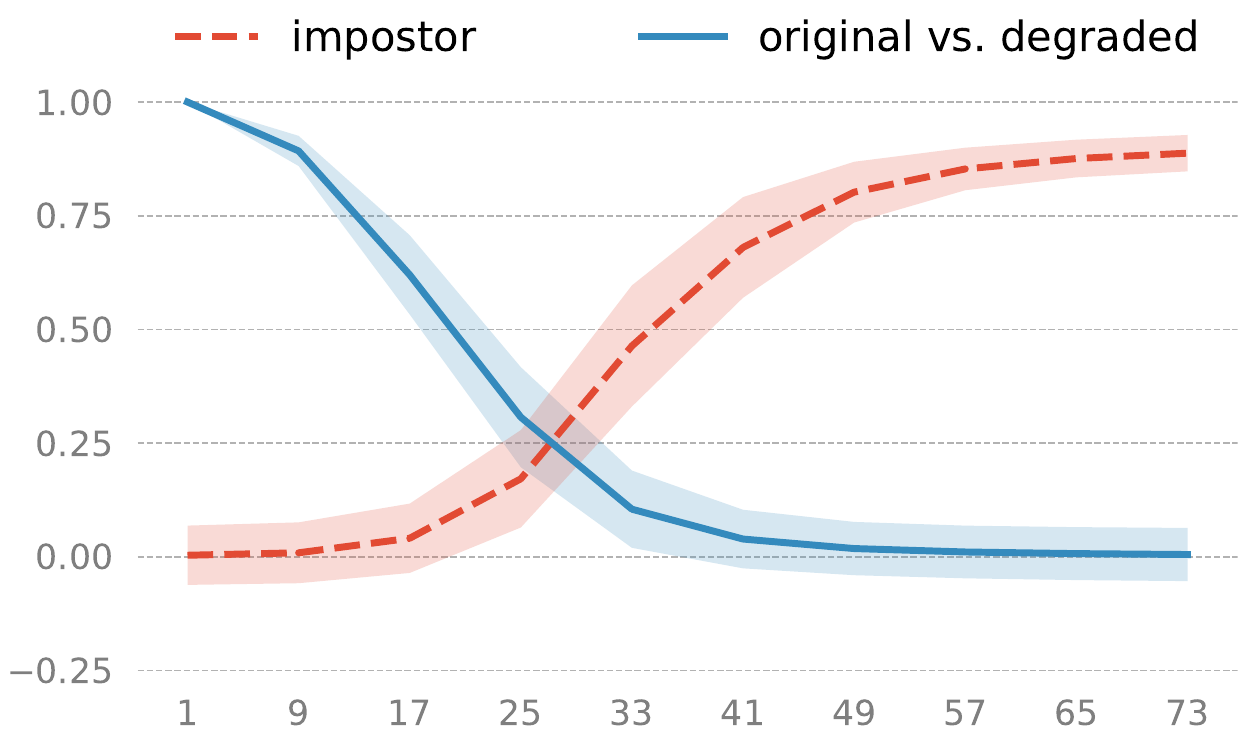}\\ %\colorbox{white}{\makebox[1em]{}}
    \centering
    \includegraphics[width=0.14\linewidth]{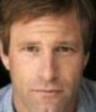}\;
    \includegraphics[width=0.14\linewidth]{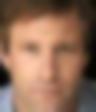}\;
    \includegraphics[width=0.14\linewidth]{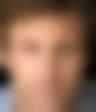}\;
    \includegraphics[width=0.14\linewidth]{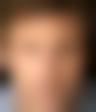}\;
    \includegraphics[width=0.14\linewidth]{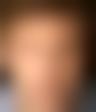}\vspace{-0.5em}
    \begin{center} \footnotesize(a) Gaussian Blur \end{center}\vspace{-1.2em}
    \end{minipage}\hfill
    \begin{minipage}{0.33\linewidth}
    \includegraphics[width=1\linewidth]{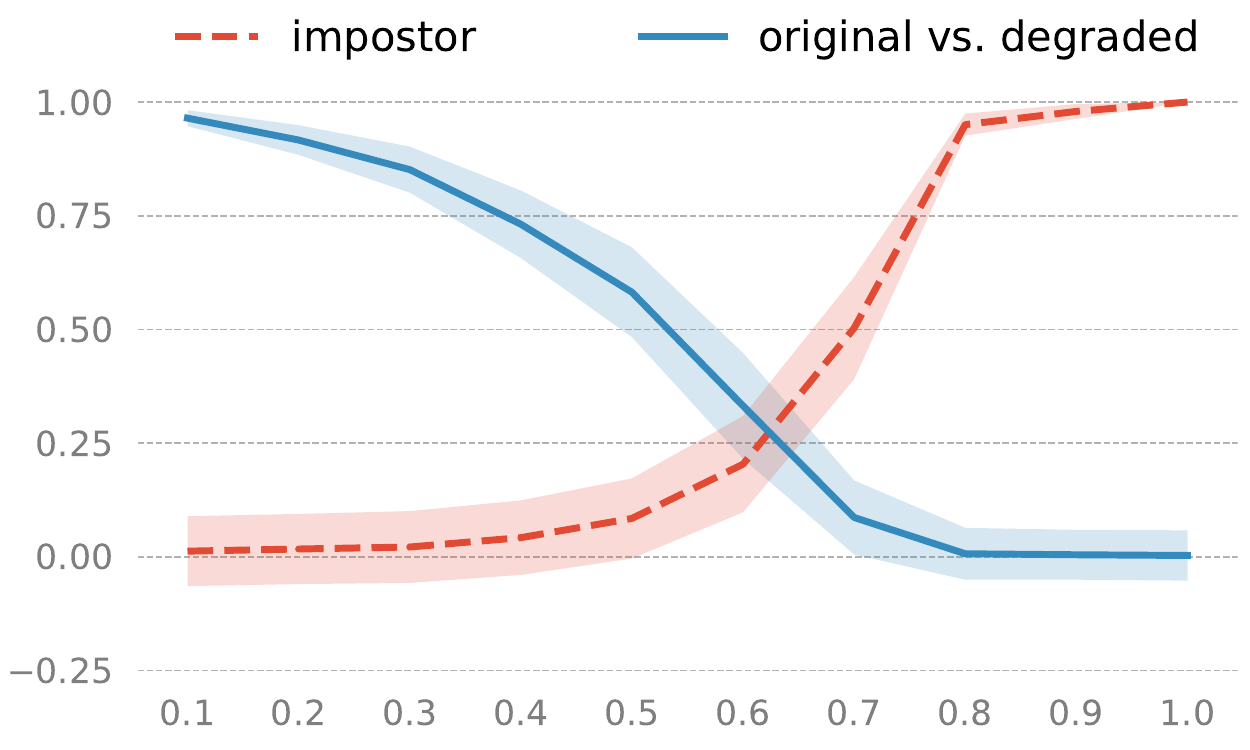}\\
    \centering
    \includegraphics[width=0.14\linewidth]{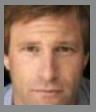}\;
    \includegraphics[width=0.14\linewidth]{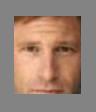}\;
    \includegraphics[width=0.14\linewidth]{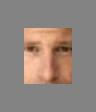}\;
    \includegraphics[width=0.14\linewidth]{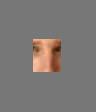}\;
    \includegraphics[width=0.14\linewidth]{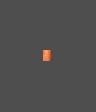}\vspace{-0.5em}
    \begin{center} \footnotesize(b) Occlusion \end{center}\vspace{-1.2em}
    \end{minipage}\hfill
    \begin{minipage}{0.33\linewidth}
    \includegraphics[width=1\linewidth]{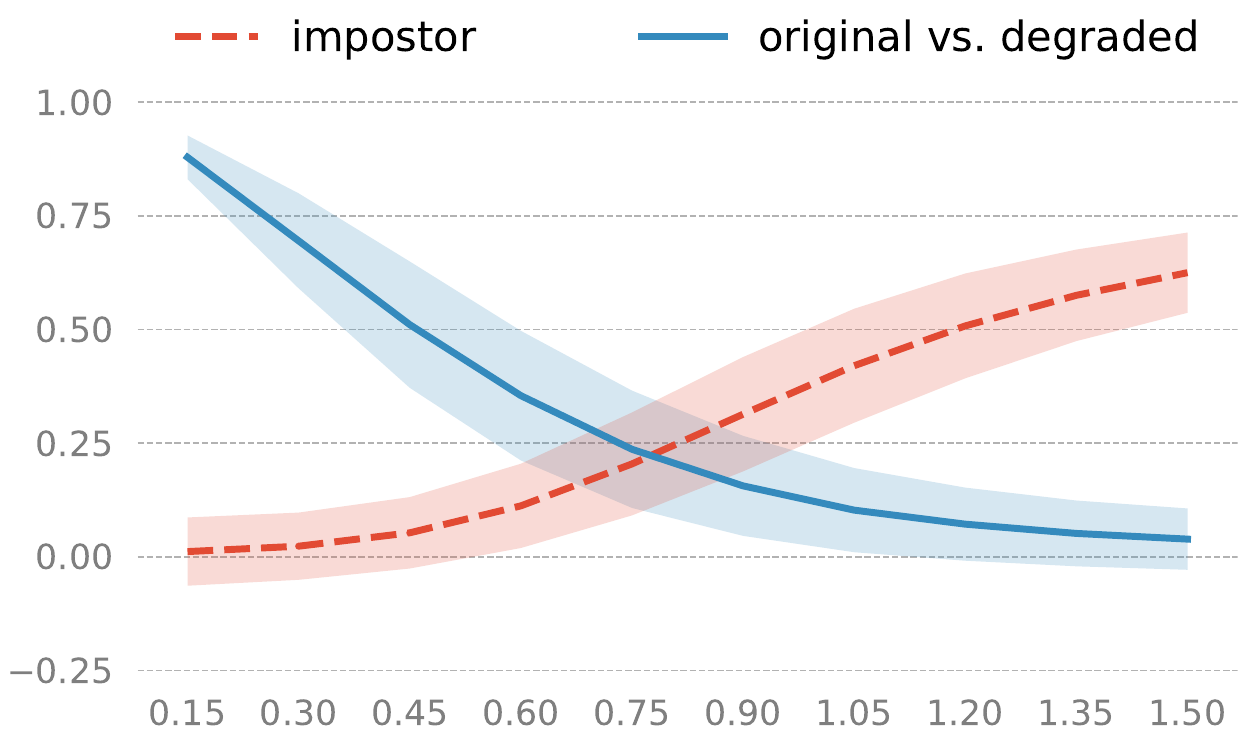}\\
    \centering
    \includegraphics[width=0.14\linewidth]{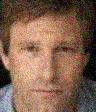}\;
    \includegraphics[width=0.14\linewidth]{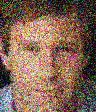}\;
    \includegraphics[width=0.14\linewidth]{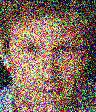}\;
    \includegraphics[width=0.14\linewidth]{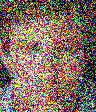}\;
    \includegraphics[width=0.14\linewidth]{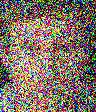}\vspace{-0.5em}
    \begin{center} \footnotesize(c) Random Gaussian Noise \end{center}\vspace{-1.2em}
    \end{minipage}\hfill
    \caption{ Illustration of \textit{feature ambiguity dilemma}. The plots show the cosine similarity on LFW dataset with different degrees of degradation. Blue lines show the similarity between original images and their respective degraded versions. Red lines show the similarity between impostor pairs of degraded images. The shading indicates the standard deviation. With larger degrees of degradation, the model becomes more confident (very high/low scores) in a wrong way.}\vspace{-1.3em}
    \label{fig:dilemma}
\end{figure*}

%-------------------------------------------------------------------------
\section{Limitations of Deterministic Embeddings}
\label{sec:motivation}

In this section, we explain the problems of deterministic face embeddings from both theoretical and empirical views. Let $\cX$ denote the image space and $\cZ$ denote the latent feature space of $D$ dimensions. An ideal latent space $\cZ$ should only encode \textit{identity-salient} features and be \textit{disentangled} from identity-irrelevant features. As such, each identity should have a unique intrinsic code $\bz\in \cZ$ that best represents this person and each face image $\bx\in\cX$ is an observation sampled from $p(\bx|\bz)$. The process of training face embeddings can be viewed as a joint process of searching for such a latent space $\cZ$ and learning the inverse mapping $p(\bz|\bx)$. For deterministic embeddings, the inverse mapping is a Dirac delta function $p(\bz|\bx)=\delta(\bz-f(\bx))$, where $f$ is the embedding function. Clearly, for any space $\cZ$, given the possibility of noises in $\bx$, it is unrealistic to recover the exact $\bz$ and the embedded point of a low-quality input would inevitably shift away from its intrinsic $\bz$ (no matter how much training data we have). 

The question is whether this shift could be bounded such that we still have smaller intra-class distances compared to inter-class distances. However, this is unrealistic for fully unconstrained face recognition and we conduct an experiment to illustrate this. Let us start with a simple example: given a pair of identical images, a deterministic embedding will always map them to the same point and therefore the distance between them will always be $0$, even if these images do not contain a face. This implies that ``a pair of images being similar or even the same does not necessarily mean the probability of their belonging to the same person is high''. 

\begin{figure}[t]
\setlength\tabcolsep{2.4px}
\newcommand{\hhh}{32px}
\newcommand{\vsp}{\hspace{0.28em}}
\newcolumntype{Y}{>{\centering\arraybackslash}X}
    \captionsetup{font=footnotesize}
    \footnotesize
    \centering
    \begin{tabularx}{\linewidth}{cc}
        \includegraphics[height=\hhh]{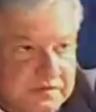}\hfill
        \includegraphics[height=\hhh]{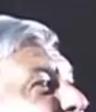}\vsp
        \includegraphics[height=\hhh]{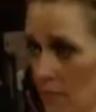}\hfill
        \includegraphics[height=\hhh]{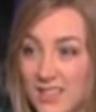} &
        \includegraphics[height=\hhh]{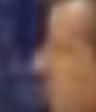}\hfill
        \includegraphics[height=\hhh]{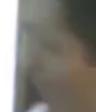}\vsp
        \includegraphics[height=\hhh]{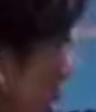}\hfill
        \includegraphics[height=\hhh]{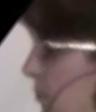} \\[-0.1em]
        \includegraphics[height=\hhh]{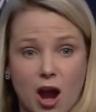}\hfill
        \includegraphics[height=\hhh]{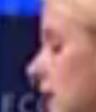}\vsp
        \includegraphics[height=\hhh]{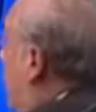}\hfill
        \includegraphics[height=\hhh]{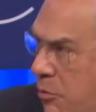} &
        \includegraphics[height=\hhh]{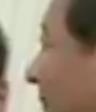}\hfill
        \includegraphics[height=\hhh]{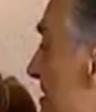}\vsp
        \includegraphics[height=\hhh]{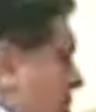}\hfill
        \includegraphics[height=\hhh]{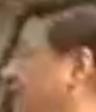} \\[-0.1em]
       (a) Low-similarity Genuine Pairs &  (b) High-similarity Impostor Pairs\\[-1.0em]
    \end{tabularx}
    \caption{Example genuine pairs from IJB-A dataset estimated with the lowest similarity scores and impostor pairs with the highest similarity scores (among all possible pairs) by a 64-layer CNN model. The genuine pairs mostly consist of one high-quality and one low-quality image while the impostor pairs are all low-quality images. Note that these pairs are not templates in the verification protocol.}\vspace{-1.5em}
    \label{fig:ijba_fail_det}
\end{figure}

To demonstrate this, we conduct an experiment by manually degrading the high-quality images and visualizing their similarity scores. We randomly select a high-quality image of each subject from the LFW dataset~\cite{LFWTech} and manually insert Gaussian blur, occlusion, and random Gaussian noise to the faces. In particular, we linearly increase the size of Gaussian kernel, occlusion ratio and the standard deviation of the noise to control the degradation degree. At each degradation level, we extract the feature vectors with a 64-layer CNN\footnote{trained on Ms-Celeb-1M~\cite{MS-CELEB} with AM-Softmax~\cite{wang2018additive}}, which is comparable to state-of-the-art face recognition systems. The features are normalized to a hyper-spherical embedding space. Then, two types of cosine similarities are reported: (1) similarity between pairs of original image and its respective degraded image, and (2) similarity between degraded images of different identities. As shown in Figure~\ref{fig:dilemma}, for all the three types of degradation, the genuine similarity scores decrease to $0$ while the impostor similarity scores converge to $1.0$! These indicate two types of errors that can be expected in a fully unconstrained  scenario even when the model is very confident (very high/low similarity scores): 
\begin{enumerate}[leftmargin=20pt,label={(\arabic*)}]\vspace{-0.5em}
    \item false accept of impostor low-quality pairs and\vspace{-0.5em}
    \item false reject of genuine cross-quality pairs.\vspace{-0.4em}
\end{enumerate}
To confirm this, we test the model on the IJB-A dataset by finding impostor/genuine image pairs with the highest/lowest scores, respectively. The situation is exactly as we hypothesized (See Figure~\ref{fig:ijba_fail_det}). We call this \textit{Feature Ambiguity Dilemma} which is observed when the deterministic embeddings are forced to estimate the features of ambiguous faces. The experiment also implies that there exist a \textit{dark space} where the ambiguous inputs are mapped to and the distance metric is distorted.

\section{Probabilistic Face Embeddings}
To address the aforementioned problem caused by data uncertainty, we propose to encode the uncertainty into the face representation and take it into account during matching. Specifically, instead of building a model that gives a point estimation in the latent space, we estimate a distribution $p(\bz|\bx)$ in the latent space to represent the potential appearance of a person's face\footnote{following the notations in Section~\ref{sec:motivation}.}. In particular, we use a multivariate Gaussian distribution:
\begin{equation}
    p(\bz|\bx_i) = \mathcal{N}(\bz;\bmu_i,\bsigma_i^{2}\mathbf{I})
\end{equation}
where $\bmu_i$ and $\bsigma_i$ are both a $D$-dimensional vector predicted by the network from the $i^{\text{th}}$ input image $\bx_i$.  Here we only consider a diagonal covariance matrix to reduce the complexity of the face representation. 
This representation should have the following properties:
\begin{enumerate}\vspace{-0.3em}
    \item The center $\bmu$ should encode the most likely facial features of the input image.\vspace{-0.6em}
    \item The uncertainty $\bsigma$ should encode the model's confidence along each feature dimension.\vspace{-0.6em}
\end{enumerate}
In addition, we wish to use a single network to predict the distribution. Considering that new approaches for training face embeddings are still being developed, we aim to develop a method that could convert existing deterministic face embedding networks to PFEs in an easy manner. In the followings, we first show how to compare and fuse the PFE representations to demonstrate their strength and then propose our method for learning PFEs.

\begin{note}
Because of the space limit, we provide the proofs of all the propositions below in Section~\ref{sec:proof}.
\end{note}

\subsection{Matching with PFEs}

% A common solution to compare between two distributions could be using a divergence,~\eg KL-divergence or Wasserstein distance. However, this is against our intuition for taking feature uncertainty into account, since the duplicate images would always have the same $\{\bmu(\bx),\bsigma(\bx)\}$ and therefore their distance is, again, always zero. Therefore, 
Given the PFE representations of a pair of images $(\bx_i,\bx_j)$, we can directly measure the ``likelihood'' of them belonging to the same person (sharing the same latent code): $p(\bz_i=\bz_j)$, where $\bz_i\sim p(\bz|\bx_i)$ and $\bz_j\sim p(\bz|\bx_j)$. Specifically,
\begin{equation}
p(\bz_i=\bz_j)=\int{p(\bz_i|\bx_i)p(\bz_j|\bx_j)\delta(\bz_i-\bz_j)d\bz_id\bz_j}.
\label{eq:likelihood_def}
\end{equation}
% \begin{proposition}
% For random variable $\Delta \bz=\bz_1-\bz_2$, it can be proved that 
% \begin{equation}
% \Delta\bz\sim\mathcal{N}(\bmu_1-\bmu_2,\diag^2{(\Tilde{\bsigma})}),
% \end{equation}
% where $\Tilde{\sigma}_j^2=\sigma_{1j}^2+\sigma_{2j}^2$ for each dimension $j$.
% \end{proposition}
% Thus, the similarity between $\bx_1$ and $\bx_2$ can be defined as
% \begin{equation}
%     s(\bx_1,\bx_2)= \log p(\bz_1=\bz_2) = \log p(\Delta \bz=\mathbf{0}),
% \end{equation}
% that is
In practice, we would like to use the log likelihood instead, whose solution is given by:
\begin{align}
\begin{split}
    s(\bx_i,\bx_j)=&\log p(\bz_i=\bz_j)\\
    =&-\frac{1}{2}\sum_{l=1}^{D}(\frac{(\mu^{(l)}_i-\mu^{(l)}_j)^2}{\sigma_i^{2(l)}+\sigma_j^{2(l)}}+\log(\sigma_i^{2(l)}+\sigma_j^{2(l)}))\\
    &-const,
\end{split}\raisetag{\baselineskip}
\label{eq:likelihood}
\end{align}
where $const=\frac{D}{2}\log{2\pi}$, $\mu^{(l)}_i$ refers to the $l^{\text{th}}$ dimension of $\bmu_i$ and similarly for $\sigma^{(l)}_i$.

Note that this symmetric measure can be viewed as the expectation of likelihood of one input's latent code conditioned on the other, that is
\begin{align}
\begin{split}
    s(\bx_i,\bx_j)=&\log\int{p(\bz|\bx_i)p(\bz|\bx_j)d\bz}\\
    =&\log\IE_{\bz\sim p(\bz|\bx_i)}[p(\bz|\bx_j)]\\
    =&\log\IE_{\bz\sim p(\bz|\bx_j)}[p(\bz|\bx_i)].
\end{split}
\end{align}
As such, we call it \textit{mutual likelihood score (MLS)}. Different from KL-divergence, this score is unbounded and cannot be seen as a distance metric. It can be shown that the squared Euclidean distance is equivalent to a special case of MLS when all the uncertainties are assumed to be the same:
\vspace{-0.3em}\begin{property}
If $\sigma_i^{(l)}$ is a fixed number for all data $\bx_i$ and dimensions $l$, MLS is equivalent to a scaled and shifted negative squared Euclidean distance.
\label{prop:euclidean}
\end{property}\vspace{-0.2em}
Further, when the uncertainties are allowed to be different, we note that MLS has some interesting properties that make it different from a distance metric:
\begin{enumerate}\vspace{-0.5em}
    \item \textit{Attention} mechanism: the first term in the bracket in Equation~(\ref{eq:likelihood}) can be seen as a weighted distance which assigns larger weights to less uncertain dimensions.\vspace{-0.5em}
    \item \textit{Penalty} mechanism: the second term in the bracket in Equation~(\ref{eq:likelihood}) can be seen as a penalty term which penalizes dimensions that have high uncertainties.\vspace{-0.5em}
    \item If either input $\bx_i$ or $\bx_j$ has large uncertainties, MLS will be low (because of penalty) irrespective of the distance between their mean.\vspace{-0.5em}
    \item Only if both inputs have small uncertainties and their means are close to each other, MLS could be very high.\vspace{-0.4em}
\end{enumerate}
The last two properties imply that PFE could solve the feature ambiguity dilemma if the network can effectively estimate $\bsigma_i$.

% \begin{proposition}
% With $(\bmu_1-\bmu_2)$ fixed, the maximum similarity is achieved when $\Tilde{\sigma}^2_{j}=(\mu_{1j}-\mu_{2j})^2$ for all dimension $j$.
% %where  \begin{equation}s_{max}=-\frac{D}{2}\sum{\log(\mu_{1i}-\mu_{2i})^2}-D-const\end{equation}
% \label{prop:similarity}
% \end{proposition}
% Based on Proposition~\ref{prop:similarity}, We can see that Equation~(\ref{eq:comparison}) has the following properties:
% \begin{enumerate}\vspace{-0.4em}
%     \item If either $\bsigma_1$ or $\bsigma_1$ is large, the similarity will be low no matter the distance between $\bmu_1$ and $\bmu_2$.\vspace{-0.5em}
%     \item Only if both $\bsigma_1$ and $\bsigma_1$ are small and $\bmu_1$ and $\bmu_2$ are close to each other, the similarity could be very high.\vspace{-0.4em}
% \end{enumerate}
% This implies that PFE could solve the feature amibiguity dilemma if the network can effectively estimate $\bsigma(\bx)$.

\begin{figure}[t]
    \centering
    \captionsetup{font=footnotesize}
    % \subfloat[example failures of embedding representation]{\includegraphics[width=0.49\linewidth]{fig/question_1.pdf}\quad
    %     \includegraphics[width=0.49\linewidth]{fig/question_2.pdf}}\\
    \quad\subfloat[]{\includegraphics[height=80pt]{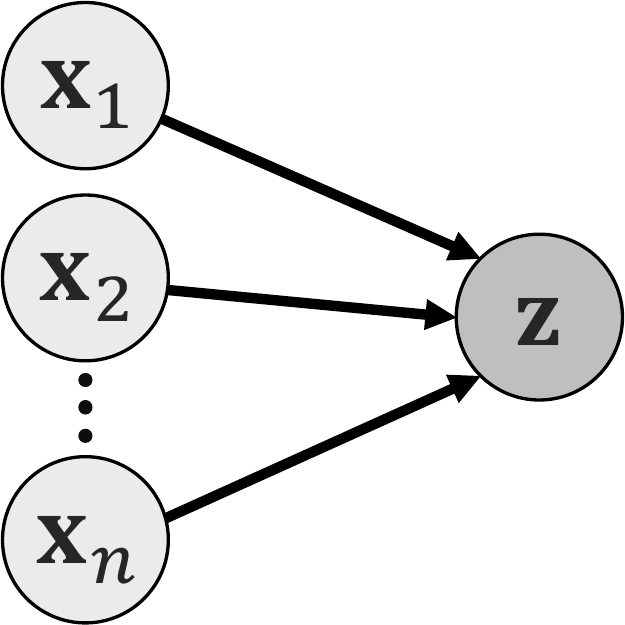}}\hfill
    \subfloat[]{\includegraphics[height=80pt]{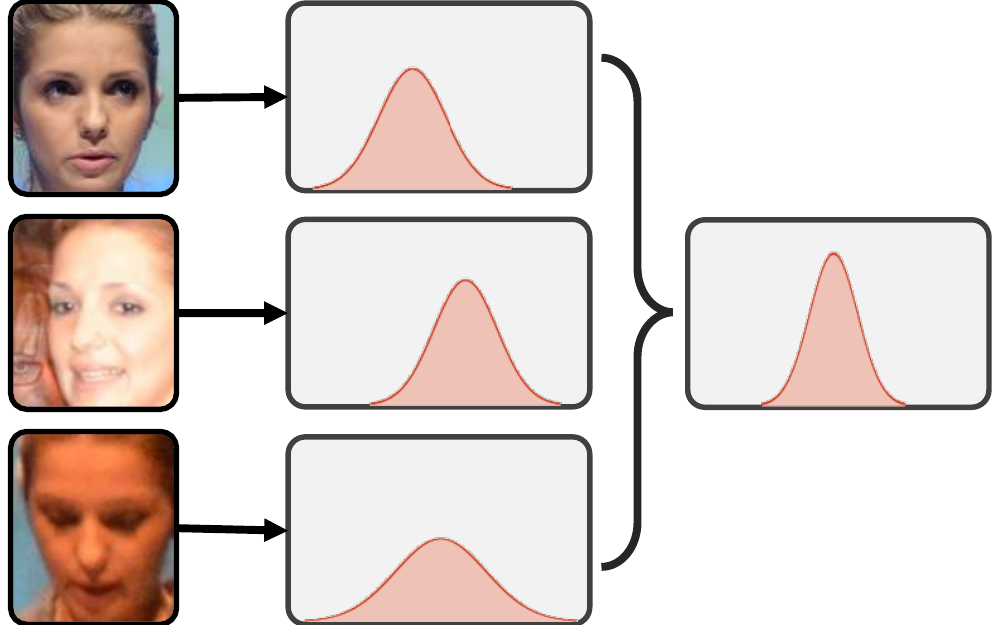}}\quad
    \vspace{-0.8em}\caption{Fusion with PFEs. (a) Illustration of the fusion process as a directed graphical model. (b) Given the Gaussian representations of faces (from the same identity), the fusion process outputs a new Gaussian distribution in the latent space with a more precise mean and lower uncertainty.}\vspace{-1.2em}
    \label{fig:fuse}
\end{figure}

\subsection{Fusion with PFEs}
In many cases we have a template (set) of face images, for which we need to build a compact representation for matching. With PFEs,
a conjugate formula can be derived for representation fusion (Figure~\ref{fig:fuse}). Let $\{\bx_1,\bx_2,\dots,\bx_{n}\}$ be a series of observations (face images) from the same identity and $p(\bz|\bx_1,\bx_2,\dots,\bx_{n})$ be the posterior distribution after the $n^{\text{th}}$ observation. Then, assuming all the observations are conditionally independent (given the latent code $\bz$). It can be shown that:
\vspace{-0.3em}\begin{align}
    p(\bz|\bx_1,\bx_2,\dots,\bx_{n+1})=\alpha\frac{p(\bz|\bx_{n+1})}{p(\bz)}p(\bz|\bx_{1},\bx_{2},\dots,\bx_{n}),
\raisetag{0.2\baselineskip}\label{eq:fusion_p}
\end{align}
% =\;& \frac{p(\bx_1,\bx_2,\dots,\bx_{i+1}|\bz)p(\bz)}{p(\bx_1,\bx_2,\dots,\bx_{i+1})}\\
% =\;& \frac{p(\bx_{i+1}|\bz)p(\bx_1,\bx_2,\dots,\bx_i|\bz)p(\bz)}{p(\bx_{i+1})p(\bx_1,\bx_2,\dots,\bx_{i})}\\
% =\;& \frac{p(\bx_{i+1}|\bz)p(\bx_1,\bx_2,\dots,\bx_i|\bz)p(\bz)}{p(\bx_{i+1})p(\bx_1,\bx_2,\dots,\bx_{i})}\\
% =\;& \frac{p(\bx_{i+1}|\bz)}{p(\bx_{i+1})}\frac{p(\bx_1,\bx_2,\dots,\bx_i|\bz)p(\bz)}{p(\bx_1,\bx_2,\dots,\bx_{i})}\\
where $\alpha$ is a normalization factor. To simplify the notations, let us only consider a one-dimensional case below; the solution can be easily extended to the multivariate case.
% Also let us define a confidence indicator $c=\frac{1}{\sigma^2}$. Let $\hat{\mu}^{(n)}$ and $\hat{c}^{(n)}$

% define a confidence vector $\bc$ to replace $\bsigma$, where $c^{(i)}_{j}=\frac{1}{\sigma_{j}^{2(i)}}$ on each dimension $j$. Also, since each dimension works in a independent way, \wlog{asdfas} let $\{\bmu_i,\bc_i\}$ denote the parameters of $p(\bz|\bx_i)$ and $\{\hat{\bmu}^{(i)},\hat{\bc}^{(i)}\}$ denote the parameters of $p(\bz|\bx_1,\bx_2,\dots,\bx_n)$.

If $p(\bz)$ is assumed to be a noninformative prior,~\ie $p(\bz)$ is a Gaussian distribution whose variance approaches $\infty$, the posterior distribution in Equation~(\ref{eq:fusion_p}) is a new Gaussian distribution with lower uncertainty (See Section~\ref{sec:proof}).
% in a one-dimensional case is a Gaussian distribution with
% \begin{gather}
% \label{eq:fusion_conjugate}
%     \hmu_{n+1}=\frac{\hsigma^2_n\mu_{n+1}+\sigma^2_{n+1}\hmu_n}{\sigma^2_{n+1}+\hsigma^2_n}, \\
%     \hsigma^2_{n+1}=\frac{\sigma^2_{n+1}\hsigma^2_n}{\sigma^2_{n+1}+\hsigma^2_n}.
% \end{gather}
Further, given a set of face images $\{\bx_1,\bx_2,\dots,\bx_{n}\}$, the parameters of the fused representation can be directly given by:
\begin{align}
    \hmu_n=\sum_{i=1}^{n}{\frac{\hsigma^2_n}{\sigma^2_i}\mu_i}, \label{eq:fusion_template_mu}\\
    \frac{1}{\hsigma^2_n}=\sum_{i=1}^{n}{\frac{1}{\sigma^2_i}}. \label{eq:fusion_template_c}
\end{align}

In practice, because the conditional independence assumption is usually not true,~\eg video frames include a large amount of redundancy, Equation~( \ref{eq:fusion_template_c}) will be biased by the number of images in the set. Therefore, we take dimension-wise minimum to obtain the new uncertainty.

\paragraph{Relationship to Quality-aware Pooling} If we consider a case where all the dimensions share the same uncertainty $\sigma_i$ for $i^{\text{th}}$ input and let the quality value $q_i=\frac{1}{\sigma^2_i}$ be the output of the network. Then Equation~(\ref{eq:fusion_template_mu}) can be written as 
\begin{equation}
    \hat{\bmu}_n=\frac{\sum_{i=1}^n{q_i\bmu_i}}{\sum_j^n q_j}.
\end{equation}
If we do not use the uncertainty after fusion, the algorithm will be the same as recent quality-aware aggregation methods for set-to-set face recognition~\cite{yang2017neural,liu2017quality,xie2018multicolumn}.

% \begin{align}
%     \bmu^{fuse}=\sum_{i=1}^{N}{\hat{\bc}_i\odot\bmu_i}, \label{eq:fusion_template_mu}\\
%     \bc^{fuse}=\sum_{i=1}^{N}{\bc_i},   \label{eq:fusion_template_c}
% \end{align}
% where
% \begin{equation}
%     \hat{\bc}_i=\bc_i\oslash\bc^{fuse}.
% \end{equation}

%-------------------------------------------------------------------------
\subsection{Learning}
% How to learn the Guassian PFE is indeed the most challenging problem in our work. On one hand, the target space of our models is a latent space where the target $\bz$ is not well-defined by given labels, so we cannot follow the data uncertainty learning in regression tasks~\cite{kendall2017uncertainties} to maximize log likelihood $\log p(\bz|\bx)$. On the other hand, although VAE~\cite{kingma2013auto} presents an unsupervised framework to learn a similar representation, its latent representation is mainly optimized for generation purposes and hence not disentangled from identity-irrelevant factors. Besides, as an approximation methods, the uncertainty in VAE is vulnerable against parameter settings and we found it not precise enough to serve our recognition task.

Note that any deterministic embedding $f$, if properly optimized, can indeed satisfy the properties of PFEs: (1) the embedding space is a disentangled identity-salient latent space and (2) $f(\bx)$ represents the most likely features of the given input in the latent space. As such, in this work we consider a stage-wise training strategy: given a pre-trained embedding model $f$, we fix its parameters, take $\bmu(\bx)=f(\bx)$, and optimize an additional uncertainty module to estimate $\bsigma(\bx)$. When the uncertainty module is trained on the same dataset of the embedding model, this stage-wise training strategy allows us to have a more fair comparison between PFE and the original embedding $f(\bx)$ than an end-to-end learning strategy.

The uncertainty module is a network with two fully-connected layers which shares the same input as of the bottleneck layer\footnote{Bottleneck layer refers to the layer which outputs the original face embedding.}. The optimization criteria is to maximize the mutual likelihood score of all genuine pairs $(\bx_i,\bx_j)$. Formally, the loss function to minimize is 
\begin{equation}
\label{eq:loss}
\EL=\frac{1}{|\mathcal{P}|}\sum_{(i,j)\in\mathcal{P}}{-s(\bx_i,\bx_j)}
\end{equation}
where $\mathcal{P}$ is the set of all genuine pairs and $s$ is defined in Equation~(\ref{eq:likelihood}). In practice, the loss function is optimized within each mini-batch. Intuitively, this loss function can be understood as an alternative to maximizing $p(\bz|\bx)$: if the latent distributions of all possible genuine pairs have a large overlap, the latent target $\bz$ should have a large likelihood $p(\bz|\bx)$ for any corresponding $\bx$. Notice that because $\bmu(\bx)$ is fixed, the optimization wouldn't lead to the collapse of all the $\bmu(\bx)$ to a single point.

% \section{Toy Example}
% In this section we conduct two toy examples to gain more insight into the proposed PFE. 

% \paragraph{MNIST} To visualize the learned PFEs, we train a variant of LeNet\footnote{\url{https://github.com/tensorflow/models/blob/master/research/slim/nets/lenet.py}} on the MNIST dataset. In order to mimic state-of-the-art face recognition systems, we first train the model with the AM-Softmax~\cite{wang2018additive} (Equivalent to CosFace~\cite{wang2018cosface}) loss function to learn a spherical embedding space. Then we fix the network and train the uncertainty module again on the same training data. The visualization is shown in Figure~\ref{???}. It can be seen that the proposed method can effectively predict the uncertainty of the samples by assigning smaller variances to samples that close to its genuine neighbors and assigning larger variances to the samples that are far away. 

% \paragraph{Feature Uncertainty Dilemma} We fix the network parameters in Section~\ref{sec:motivation} and train an extra uncertainty module on the same training data to output the variance estimation. Then we repeat the same experiments in Section~\ref{sec:motivation} by using the Equation~\ref{eq:comparison} as the score metric. As shown in Figure~\ref{fig:???}, although the impostor scores still increase with more degradation, they never surpass the genuine ones. Instead they converge to a point in between. The results shows that the proposed PFE can effectively solve the feature ambiguity dilemma, which could lead to potential improvement in face recognition performance.

\section{Experiments}
In this section, we first test the proposed PFE method on standard face recognition protocols to compare with deterministic embeddings. Then we conduct qualitative analysis to gain more insight into how PFE behaves. \textit{Due to the space limit, we provide the implementation details in Section~\ref{sec:detail}.}

% \subsection{Datasets}
To comprehensively evaluate the efficacy of PFEs, we conduct the experiments on $7$ benchmarks, including the well known \textbf{LFW}~\cite{LFWTech}, \textbf{YTF}~\cite{YTF},  \textbf{MegaFace}~\cite{kemelmacher2016megaface} and four other more unconstrained benchmarks:
% \\\textbf{LFW}~\cite{LFWTech} contains $13,233$ near-frontal and high-quality face photos of $5,749$ subjects. The verification protocol used in this paper includes $6000$ face pairs.
% \\\textbf{YTF}~\cite{YTF} contains $3,425$ videos $1,595$ subjects. The verification protocol used in this paper includes $5,000$ video pairs. 
% \\\textbf{MegaFace}~\cite{kemelmacher2016megaface} contains 1M face images from Flicker as distractors. The FaceScrub dataset is used as the probe set in our experiments, which contains $3530$ high-quality face images of $80$ subjects.
\\\textbf{CFP}~\cite{CFP} contains $7,000$ frontal/profile face photos of $500$ subjects. We only test on the frontal-profile (FP) protocol, which includes $7,000$ pairs of frontal-profile faces.
\\\textbf{IJB-A}~\cite{IJBA} is a template-based benchmark, containing $25,813$ faces images of $500$ subjects. Each template includes a set of still photos or video frames. Compared with previous benchmarks, the faces in IJB-A have larger variations and present a more unconstrained scenario.
\\\textbf{IJB-C}~\cite{IJBC} is an extension of IJB-A with $140,740$ faces images of $3,531$ subjects. The verification protocol of IJB-C includes more impostor pairs so we can compute True Accept Rates (TAR) at lower False Accept Rates (FAR).
\\\textbf{IJB-S}~\cite{IJBS} is a surveillance video benchmark containing $350$ surveillance videos spanning $30$
hours in total, $5,656$ enrollment images, and $202$ enrollment videos of $202$ subjects. Many faces in this dataset are of extreme pose or low-quality, making it one of the most challenging face recognition benchmarks (See Figure~\ref{fig:dataset} for example images).

We use the CASIA-WebFace~\cite{CASIA-WebFace} and a cleaned version\footnote{\url{https://github.com/inlmouse/MS-Celeb-1M_WashList.}} of MS-Celeb-1M~\cite{MS-CELEB} as training data, from which we remove the subjects that are also included in the test datasets\footnote{$84$ and $4,182$ subjects were removed from CASIA-WebFace and MS-Celeb-1M, respectively.}.

\begin{table}[t]
\captionsetup{font=footnotesize}
\footnotesize
\setlength{\tabcolsep}{3pt}
\begin{center}
\begin{tabularx}{\linewidth}{Xccccc}
\toprule
Base Model                      & Representation & LFW & YTF & CFP-FP & IJB-A \\
\midrule
                                                        & Original  & $98.93$ & $94.74$  & $93.84$ & $78.16$\\
\multirow{-2}{*}{\shortstack[l]{Softmax + \\Center Loss}~\cite{wen2016discriminative}}  
                                                        & PFE       & $\bold{99.27}$ & $\bold{95.42}$ & $\bold{94.51}$ & $\bold{80.83}$ \\\rowcolor{Gray}
%%%%%%%%%%%%%%%%%%%%%%%%%%%%%%%%%%%%%%%%%%%%%%%%%%%%%%%%%%%%%%%%%%%%%%%%%%%%%%%%%%%%%%%%%%%%%%%%%%%%%%%%%%%%%%
                                                        & Original  & $97.65$ & $93.36$ & $89.76$ & $60.82$ \\\rowcolor{Gray}
\multirow{-2}{*}{Triplet~\cite{schroff2015facenet}}     & PFE       & $\bold{98.45}$ & $\bold{93.96}$ & $\bold{90.04}$ & $\bold{61.00}$ \\
%%%%%%%%%%%%%%%%%%%%%%%%%%%%%%%%%%%%%%%%%%%%%%%%%%%%%%%%%%%%%%%%%%%%%%%%%%%%%%%%%%%%%%%%%%%%%%%%%%%%%%%%%%%%%%
                                                        & Original  & $99.15$ & $94.80$ & $92.41$ & $78.54$ \\
\multirow{-2}{*}{A-Softmax~\cite{liu2017sphereface}}      & PFE       & $\bold{99.32}$ & $\bold{94.94}$ & $\bold{93.37}$ & $\bold{82.58}$ \\\rowcolor{Gray}
%%%%%%%%%%%%%%%%%%%%%%%%%%%%%%%%%%%%%%%%%%%%%%%%%%%%%%%%%%%%%%%%%%%%%%%%%%%%%%%%%%%%%%%%%%%%%%%%%%%%%%%%%%%%%%
                                                        & Original  & $99.28$ & $95.64$ & $94.77$ & $84.69$ \\\rowcolor{Gray}
\multirow{-2}{*}{AM-Softmax~\cite{wang2018additive}}    & PFE       & $\bold{99.55}$ & $\bold{95.92}$ & $\bold{95.06}$ & $\bold{87.58}$ \\
%%%%%%%%%%%%%%%%%%%%%%%%%%%%%%%%%%%%%%%%%%%%%%%%%%%%%%%%%%%%%%%%%%%%%%%%%%%%%%%%%%%%%%%%%%%%%%%%%%%%%%%%%%%%%%
%                                                         & Original  & $99.42$ & $-$ & $88.96$ & $-$ \\\rowcolor{Gray}
% \multirow{-2}{*}{A-Softmax~\cite{deng2018arcface}}      & PFE       & $99.55$ & $-$ & $91.01$ & $-$ \\
\bottomrule
\end{tabularx}
\vspace{-0.9em}\caption{Results of models trained on CASIA-WebFace. ``Original'' refers to the deterministic embeddings. The better performance among each base model are shown in bold numbers. ``PFE'' uses mutual likelihood score for matching. IJB-A results are verification rates at FAR=$0.1\%$.}\vspace{-2.5em}
\label{tab:loss_functions}
\end{center}
\end{table}

\subsection{Experiments on Different Base Embeddings}\vspace{-0.3em}
Since our method works by converting existing deterministic embeddings, we want to evaluate how it works with different base embeddings,~\ie face representations trained with different loss functions. In particular, we implement the following state-of-the-art loss functions: Softmax+Center Loss~\cite{wen2016discriminative}, Triplet Loss~\cite{schroff2015facenet}, A-Softmax~\cite{liu2017sphereface} and AM-Softmax~\cite{wang2018additive}\footnote{We also tried implementing ArcFace~\cite{deng2018arcface} but it does not converge well in our case. So we did not use it.}. To be aligned with previous work~\cite{liu2017sphereface,wang2018cosface}, we train a 64-layer residual network~\cite{liu2017sphereface} with each of these loss functions on the CASIA-WebFace dataset as base models. All the features are $\ell2$-normalized to a hyper-spherical embedding space. Then we train the uncertainty module for each base model on the CASIA-WebFace again for $3,000$ steps. We evaluate the performance on four benchmarks: LFW~\cite{LFWTech}, YTF~\cite{YTF}, CFP-FP~\cite{CFP} and IJB-A~\cite{IJBA}, which present different challenges in face recognition. The results are shown in Table~\ref{tab:loss_functions}. The PFE improves over the original representation in all cases, indicating the proposed method is robust with different embeddings and testing scenarios.

\begin{table}[t]
\captionsetup{font=footnotesize}
\newcommand{\mr}[1]{\multirow{2}{*}{#1}}
\setlength{\tabcolsep}{4pt}
\footnotesize
\begin{center}
\resizebox{1.0\columnwidth}{!}{%
\begin{tabularx}{1.0\linewidth}{Xccccc}
\toprule
\mr{Method}      & \mr{Training Data}     & \mr{LFW} & \mr{YTF} & MF1    & MF1 \\
            &                   &     &     & Rank1  & Veri.            \\
\midrule            
DeepFace+~\cite{taigman2014deepface}    & 4M    & $97.35$   & $91.4$    & -  & -        \\
FaceNet~\cite{schroff2015facenet}       & 200M  & $99.63$   & $95.1$    & -  & -        \\
DeepID2+~\cite{deepid2plus}             & 300K  & $99.47$   & $93.2$    & -  & -        \\
CenterFace~\cite{wen2016discriminative} & 0.7M  & $99.28$   & $94.9$    & $65.23$ & $76.52$    \\
SphereFace~\cite{liu2017sphereface}     & 0.5M  & $99.42$   & $95.0$    & $75.77$ & $89.14$   \\
ArcFace~\cite{deng2018arcface}          & 5.8M  & $99.83$  & $98.02$    & $81.03$   & $96.98$    \\
CosFace~\cite{wang2018cosface}          & 5M    & $99.73$   & $97.6$    & $77.11$ & $89.88$    \\
L2-Face~\cite{ranjan2017l2}             & 3.7M  & $99.78$   & $96.08$   & -  & -        \\\hline
Baseline                                & 4.4M  & $99.70$   & $97.18$   & $79.43$   & $92.93$   \\
PFE\textsubscript{fuse}                 & 4.4M  & -         & $97.32$   & -  & -        \\
PFE\textsubscript{fuse+match}           & 4.4M  & ${99.82}$ & $97.36$   & $78.95$ & $92.51$    \\
\bottomrule
\end{tabularx}
}
\vspace{-0.9em}\caption{Results of our models (last three rows) trained on MS-Celeb-1M and state-of-the-art methods on LFW, YTF and MegaFace. The MegaFace verification rates are computed at FAR=$0.0001\%$. ``-'' indicates that the author did report the performance on the corresponding protocol.}\vspace{-2.8em}
\label{tab:lfw}
\end{center}
\end{table}

\begin{table}[t]
\captionsetup{font=footnotesize}
\newcommand{\mr}[1]{\multirow{2}{*}{#1}}
\footnotesize
\setlength{\tabcolsep}{1.1pt}
\begin{center}
\begin{tabularx}{1.00\linewidth}{Xc c cc c c}
\toprule
\mr{Method}                             & \mr{Training Data} &\;& \multicolumn{2}{c}{IJB-A (TAR@FAR)} &\;& \mr{\;CFP-FP} \\
\cline{4-5}\\[-1.0em]
                                        &           &&           $0.1\%$      & $1.0\%$   &&         \\
\midrule            
DR-GAN~\cite{tran2017disentangled}      & 1M        && $53.9\pm4.3$  & $77.4\pm2.7$ &&$93.41$         \\
Yin~\etal~\cite{yin2018multi}           & 0.5M      && $73.9\pm4.2$  & $77.5\pm2.5$ &&$\mathbf{94.39}$         \\
TPE~\cite{sankaranarayanan2016triplet}  & 0.5M      && $90.0\pm1.0$  & $93.4\pm0.5$ &&$89.17$         \\
NAN~\cite{yang2017neural}               & 3M        && $88.1\pm1.1$  & $94.1\pm0.8$ &&-               \\
QAN~\cite{liu2017quality}               & 5M        && $89.31\pm3.92$  & $94.20\pm1.53$  &&-              \\
Cao~\etal~\cite{cao2018vggface2}        & 3.3M      && $90.4\pm1.4$  & $95.8\pm0.6$ &&-               \\
Multicolumn~\cite{xie2018multicolumn}   & 3.3M      && $92.0\pm1.3$  & $96.2\pm0.5$ &&-               \\
L2-Face~\cite{ranjan2017l2}             & 3.7M      && $94.3\pm0.5$  & $97.00\pm0.4$&&-               \\\hline
Baseline                                & 4.4M      && $93.30\pm1.29$& $96.15\pm0.71$ &&$92.78$        \\
PFE\textsubscript{fuse}                 & 4.4M      && $94.59\pm0.72$& $95.92\pm0.73$ &&-             \\
PFE\textsubscript{fuse+match}           & 4.4M      && $\mathbf{95.25\pm0.89}$ & $\mathbf{97.50\pm0.43}$ &&$93.34$    \\
\bottomrule
\end{tabularx}
\vspace{-0.7em}\caption{Results of our models (last three rows) trained on MS-Celeb-1M  and state-of-the-art methods on CFP (frontal-profile protocol) and IJB-A.}\vspace{-1.5em}
\label{tab:ijb}
\end{center}
\end{table}
\begin{table}[t]
\newcommand{\hly}{\cellcolor{Y}}
\newcommand{\hlg}{\cellcolor{G}}
\captionsetup{font=footnotesize}
\newcommand{\mr}[1]{\multirow{2}{*}{#1}}
\footnotesize
\setlength{\tabcolsep}{4.6pt}
\begin{center}
\begin{tabularx}{1.00\linewidth}{Xc cccc}
\toprule
\mr{Method}                                  & \mr{Training Data} & \multicolumn{4}{c}{IJB-C (TAR@FAR)} \\
\cline{3-6}\\[-1.0em]
                                        &           & $0.001\%$   & $0.01\%$  & $0.1\%$ & $1\%$\\
\midrule            
Yin~\etal~\cite{yin2018towards}         & 0.5M      & - & - & $69.3$ & $83.8$\\
Cao~\etal~\cite{cao2018vggface2}        & 3.3M      & $74.7$ & $84.0$ & $91.0$ & $96.0$\\
Multicolumn~\cite{xie2018multicolumn}   & 3.3M      & $77.1$ & $86.2$ & $92.7$ & $96.8$\\
DCN~\cite{xie2018comparator}            & 3.3M      & - & $88.5$ & $94.7$ & $\mathbf{98.3}$\\\hline
Baseline                                & 4.4M      & $70.10$ & $85.37$   & $93.61$ & $96.91$\\
PFE\textsubscript{fuse}                 & 4.4M      & $83.14$ & $92.38$   & $95.47$ & $97.36$\\
PFE\textsubscript{fuse+match}           & 4.4M      & $\mathbf{89.64}$ & $\mathbf{93.25}$   & $\mathbf{95.49}$ & $97.17$\\
\bottomrule
\end{tabularx}
\vspace{-0.7em}\caption{Results of our models (last three rows) trained on MS-Celeb-1M and state-of-the-art methods on IJB-C.}\vspace{-2.8em}
\label{tab:ijbc}
\end{center}
\end{table}

\subsection{Comparison with State-Of-The-Art}
\label{sec:exp:msceleb}

\begin{table*}[!h]
\newcommand{\hly}{\cellcolor{Y}}
\newcommand{\hlg}{\cellcolor{G}}
\captionsetup{font=footnotesize}
    \centering
    \footnotesize
    \setlength{\tabcolsep}{1.5pt}
		\centering
        \scalebox{1.0}{
		\begin{tabularx}{\linewidth}{Xc c ccccc c ccccc c ccccc}
		\toprule
		\multirow{2}{*}{Method} & \multirow{2}{*}{Training Data} && \multicolumn{5}{c}{Surveillance-to-Single} && \multicolumn{5}{c}{Surveillance-to-Booking} && \multicolumn{5}{c}{Surveillance-to-Surveillance} \\
		\cline{4-8} \cline{10-14} \cline{16-20}
		&               && Rank-1 & Rank-5 & Rank-10 & 1\%  & 10\%   
		                && Rank-1 & Rank-5 & Rank-10 & 1\%  & 10\%   
		                &&  Rank-1 & Rank-5 & Rank-10 & 1\%  & 10\% \\
		\midrule
		C-FAN~\cite{gong2019video}    & 5.0M   
		                && $50.82$ & $61.16$ & $64.95$ & $16.44$ & $24.19$ 
		                && $53.04$ & $62.67$ & $66.35$ & $27.40$ & $29.70$ 
		                && $\mathbf{10.05}$ & $17.55$ & $21.06$ & $0.11$ & $0.68$\\
	    Baseline        & 4.4M  
	                    && $50.00$ & $59.07$ & $62.70$ & $7.22$ & $19.05$ 
		                && $47.54$ & $56.14$ & $61.08$ & $14.75$ & $22.99$ 
		                && $9.40$ & $17.52$ & $23.04$ & $0.06$ & $0.71$\\
		PFE\textsubscript{fuse}  & 4.4M  
		                && $\mathbf{53.44}$ & $\mathbf{61.40}$ & $\mathbf{65.05}$ & $10.53$ & $22.87$ 
		                && $\mathbf{55.45}$ & $\mathbf{63.17}$ & $\mathbf{66.38}$ & $16.70$ & $26.20$ 
		                && $8.18$ & $14.52$ & $19.31$ & $0.09$ & $0.63$\\
	    PFE\textsubscript{fuse+match}    & 4.4M  
	                    && $50.16$ & $58.33$ & $62.28$ & $\mathbf{31.88}$ & $\mathbf{35.33}$ 
		                && $53.60$ & $61.75$ & $64.97$ & $\mathbf{35.99}$ & $\mathbf{39.82}$ 
		                && $9.20$ & $\mathbf{20.82}$ & $\mathbf{27.34}$ & $\mathbf{0.84}$ & $\mathbf{2.83}$\\
		\bottomrule
		\end{tabularx}}
    \vspace{-0.7em}\caption{Performance comparison on three protocols of IJB-S. The performance is reported in terms of rank retrieval (closed-set) and TPIR@FPIR (open-set) instead of the media-normalized version~\cite{IJBS}. The numbers ``$1\%$'' and ``$10\%$'' in the second row refer to the FPIR.}\vspace{-0.5em}
    \label{tab:ijbs}
\end{table*}

\begin{figure*}[t]
    \centering
    \captionsetup{font=footnotesize}
    \begin{minipage}{0.33\linewidth}
    \includegraphics[width=1\linewidth]{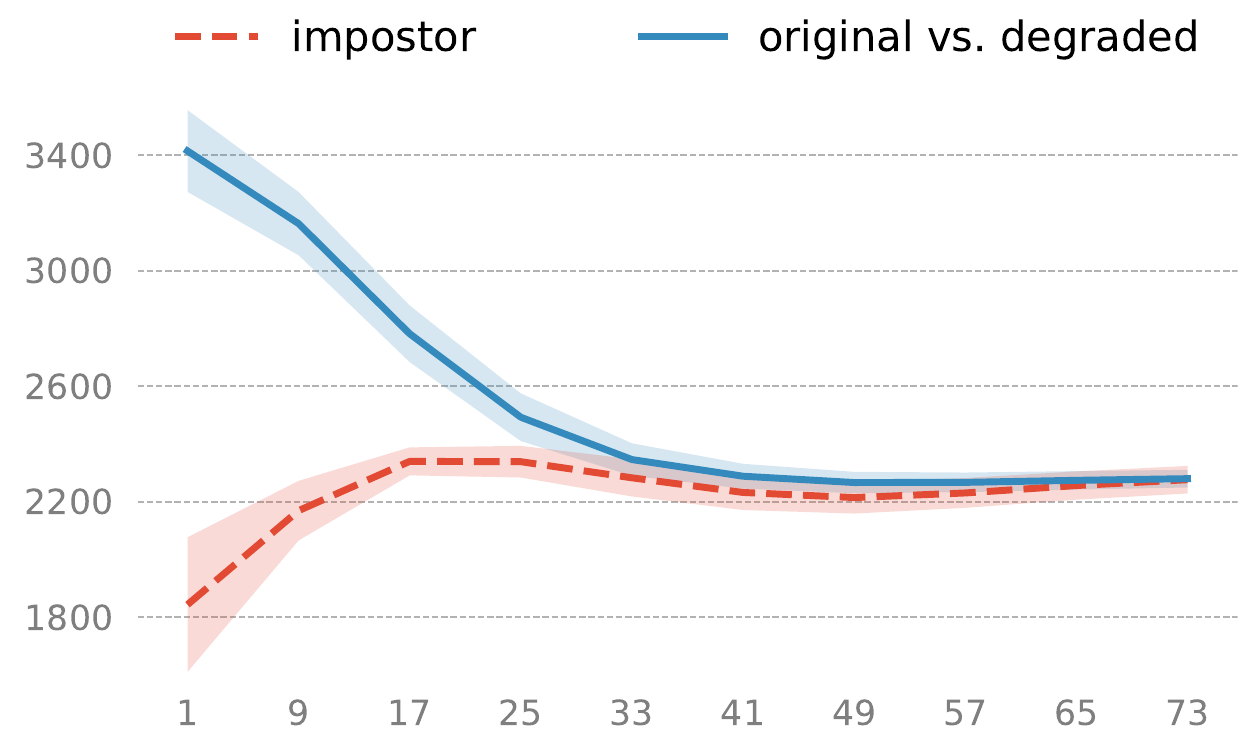}\\[-2.0em]
    \begin{center} \footnotesize(a) Gaussian Blur \end{center}\vspace{-1.2em}
    \end{minipage}\hfill
    \begin{minipage}{0.33\linewidth}
    \includegraphics[width=1\linewidth]{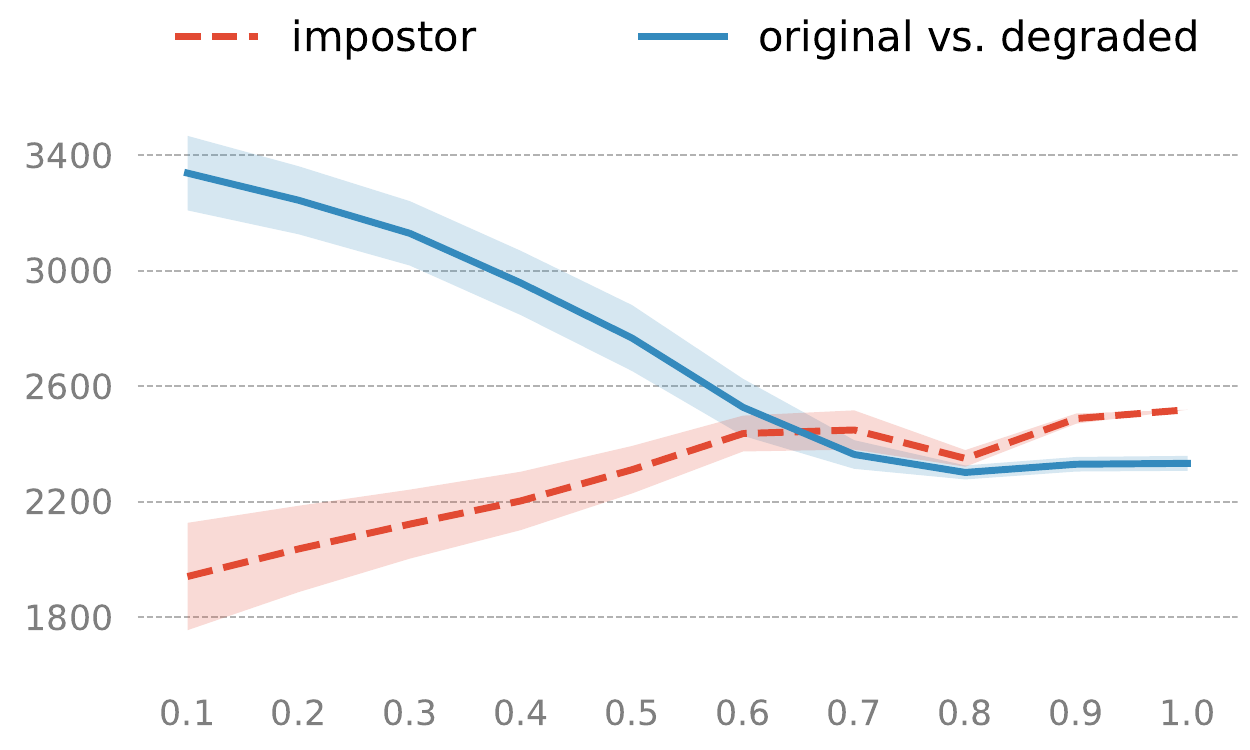}\\[-2.0em]
    \begin{center} \footnotesize(b) Occlusion \end{center}\vspace{-1.2em}
    \end{minipage}\hfill
    \begin{minipage}{0.33\linewidth}
    \includegraphics[width=1\linewidth]{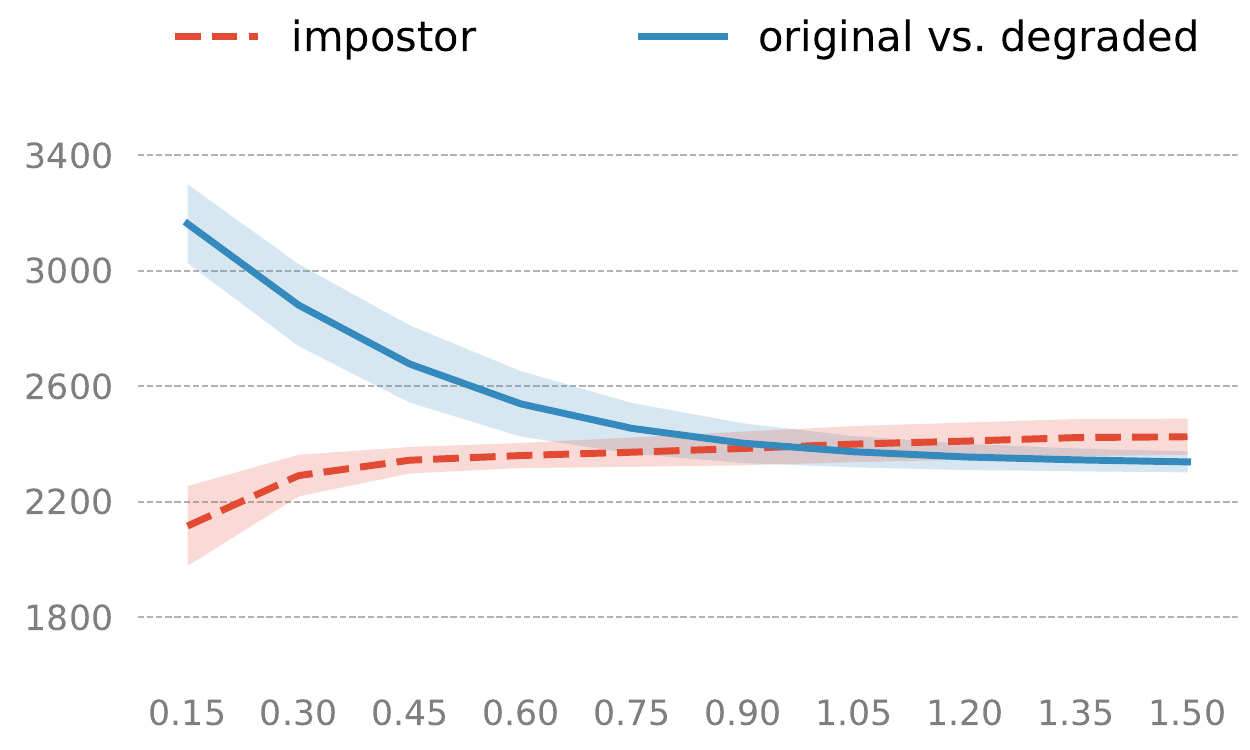}\\[-2.0em]
    \begin{center} \footnotesize(c) Random Noise \end{center}\vspace{-1.2em}
    \end{minipage}\hfill
    \vspace{-0.3em}\caption{ Repeated experiments on feature ambiguity dilemma with the proposed PFE. The same model in Figure~\ref{fig:dilemma} is used as the base model and is converted to a PFE by training an uncertainty module. No additional training data nor data augmentation is used for training.}\vspace{-1.0em}
    \label{fig:dilemma_repeat}
\end{figure*}
\begin{figure}[t]
\setlength\tabcolsep{2.4px}
\newcommand{\hhh}{32px}
\newcommand{\vsp}{\hspace{0.28em}}
\newcolumntype{Y}{>{\centering\arraybackslash}X}
    \captionsetup{font=footnotesize}
    \footnotesize
    \centering
    \begin{tabularx}{\linewidth}{cc}
        \includegraphics[height=\hhh]{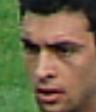}\hfill
        \includegraphics[height=\hhh]{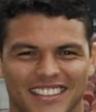}\vsp
        \includegraphics[height=\hhh]{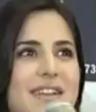}\hfill
        \includegraphics[height=\hhh]{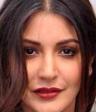} &
        \includegraphics[height=\hhh]{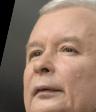}\hfill
        \includegraphics[height=\hhh]{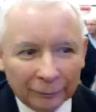}\vsp
        \includegraphics[height=\hhh]{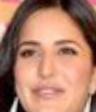}\hfill
        \includegraphics[height=\hhh]{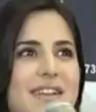} \\[-0.1em]
        \includegraphics[height=\hhh]{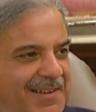}\hfill
        \includegraphics[height=\hhh]{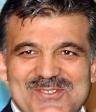}\vsp
        \includegraphics[height=\hhh]{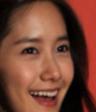}\hfill
        \includegraphics[height=\hhh]{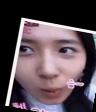} &
        \includegraphics[height=\hhh]{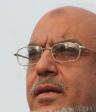}\hfill
        \includegraphics[height=\hhh]{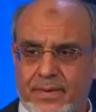}\vsp
        \includegraphics[height=\hhh]{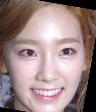}\hfill
        \includegraphics[height=\hhh]{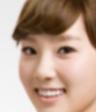} \\[-0.1em]
       (a) Low-score Genuine Pairs &  (b) High-score Impostor Pairs\\[-1.0em]
    \end{tabularx}
    \caption{Example genuine pairs  from IJB-A dataset estimated with the lowest mutual likelihood scores and impostor pairs with the highest scores by the PFE version of the same 64-layer CNN model in Section~\ref{sec:motivation}. In comparison to Figure~\ref{fig:ijba_fail_det}, most images here are high-quality ones with clear features, which can mislead the model to be confident in a wrong way. Note that these pairs are not templates in the verification protocol.}\vspace{-1.5em}
    \label{fig:ijba_fail_prob}
\end{figure}

To compare with state-of-the-art face recognition methods, we use a different base model, which is a $64$-layer network trained with AM-Softmax on the MS-Celeb-1M dataset. Then we fix the parameters and train the uncertainty module on the same dataset for $12,000$ steps. In the following experiments, we compare 3 methods:
\begin{itemize}[leftmargin=*]\vspace{-0.5em}
    \item \textbf{Baseline} only uses the original features of the 64-layer deterministic embedding along with cosine similarity for matching. Average pooling is used in case of template/video benchmarks.\vspace{-0.6em}
    \item \textbf{PFE\textsubscript{fuse}} uses the uncertainty estimation $\bsigma$ in PFE and Equation~(\ref{eq:fusion_template_mu}) to aggregate the features of templates but uses cosine similarity for matching. If the uncertainty module could estimate the feature uncertainty effectively, fusion with $\bsigma$ should be able to outperform average pooling by assigning larger weights to confident features.\vspace{-0.5em}
    \item \textbf{PFE\textsubscript{fuse+match}} uses $\bsigma$ both for fusion and matching (with mutual likelihood scores). Templates/videos are fused based on Equation~(\ref{eq:fusion_template_mu}) and Equation~(\ref{eq:fusion_template_c}).\vspace{-0.2em}
\end{itemize}

In Table~\ref{tab:lfw} we show the results on three relatively easier benchmarks: LFW, YTF and MegaFace. Although the accuracy on LFW and YTF are nearly saturated, the proposed PFE still improves the performance of the original representation. Note that MegaFace is a biased dataset: because all the probes are high-quality images from FaceScrub, the positive pairs in MegaFace are both high-quality images while the negative pairs only contain at most one low-quality image\footnote{The negative pairs of MegaFace in the verification protocol only include those between probes and distractors.} . Therefore, neither of the two types of error caused by the feature ambiguity dilemma (Section~\ref{sec:motivation}) will show up in MegaFace and it naturally favors deterministic embeddings. However, the PFE still maintains the performance in this case. We also note that such a bias, namely the target gallery images being of higher quality than the rest of gallery, would not exist in real world applications.

In Table~\ref{tab:ijb} and Table~\ref{tab:ijbc} we show the results on three more challenging datasets: CFP, IJB-A and IJB-C. The images in these datasets present larger variations in pose, occlustion, etc, and facial features could be more ambiguous. As such, we can see that PFE achieves a more significant improvement on these three benchmarks. In particular on IJB-C at FAR$=0.001\%$, PFE reduces the error rate by $64\%$. In addition, simply fusing the original features with the learned uncertainty (PFE\textsubscript{fuse}) also helps the performance. 

In Table~\ref{tab:ijbs} we report the results on three protocols of the latest benchmark, IJB-S. Again, PFE is able to improve the performance in most cases.
Notice that the gallery templates in the ``Surveillance-to-still'' and ``Surveillance-to-booking'' all include high-quality frontal mugshots, which present little feature ambiguity. Therefore, we only see a slight performance gap in these two protocols. But in the most challenging ``surveillance-to-surveillance'' protocol, larger improvement can be achieved by using uncertainty for matching. Besides, PFE\textsubscript{fuse+match} improves the performance significantly on all the open-set protocols, which indicates that MLS has more impact on the absolute pairwise score than the relative ranking.

\begin{figure}[t]
    \centering
    \captionsetup{font=footnotesize}
    \vspace{-0.5em}\includegraphics[width=0.8\linewidth]{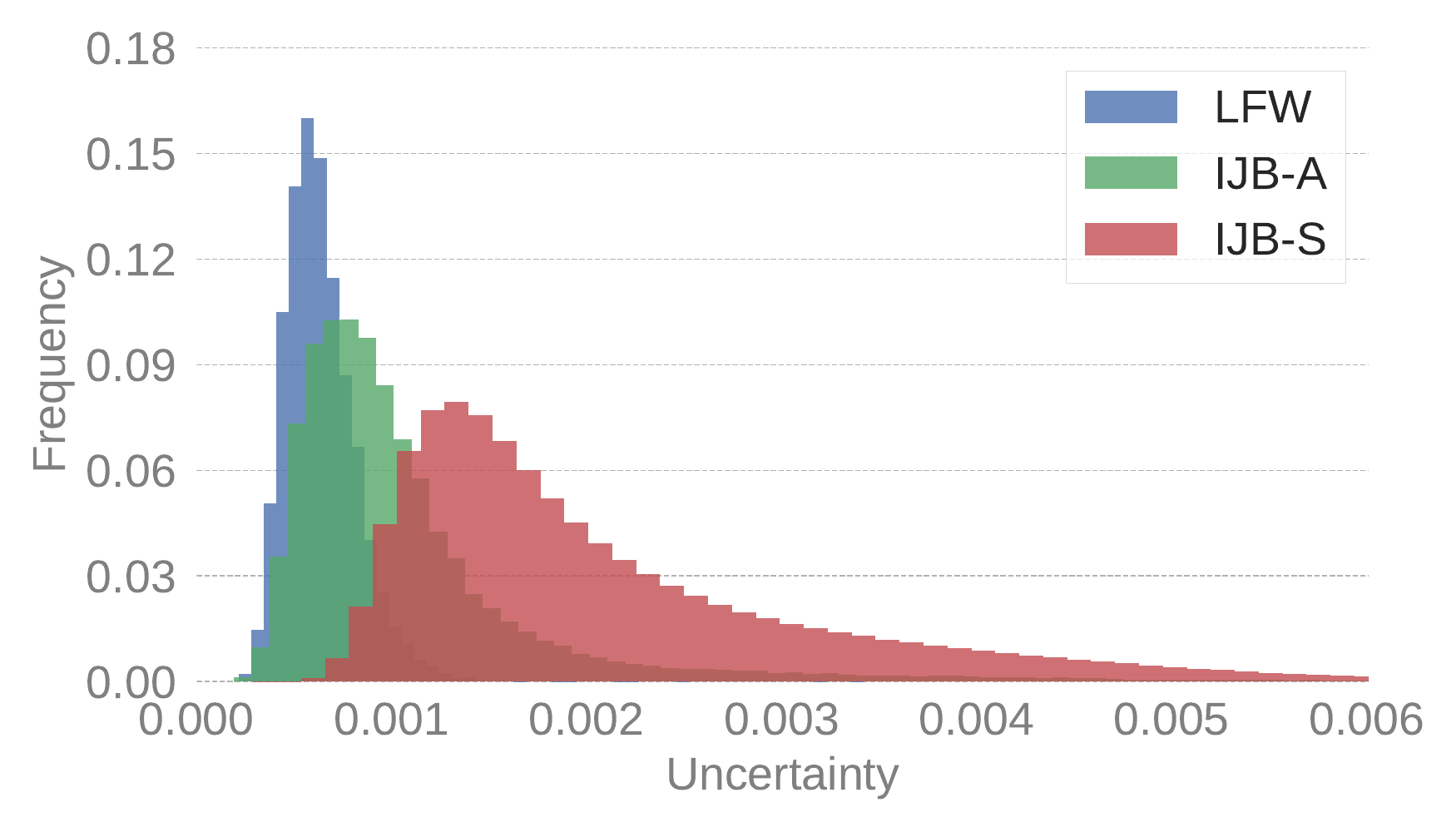}\\[-1.0em]
    \vspace{-0.2em}\caption{ Distribution of estimated uncertainty on different datasets. Here, ``Uncertainty'' refers to the harmonic mean of $\sigma$ across all feature dimensions. Note that the estimated uncertainty is proportional to the complexity of the datasets. \textbf{Best viewed in color}.}\vspace{-1.5em}
    \label{fig:hist_sigma}
\end{figure}

\subsection{Qualitative Analysis}

\paragraph{Why and when does PFE improve performance?} We first repeat the same experiments in Section~\ref{sec:motivation} using the PFE representation and MLS. The same network is used as the base model here. As one can see in Figure~\ref{fig:dilemma_repeat}, although the scores of low-quality impostor pairs are still increasing, they converge to a point that is lower than the majority of genuine scores. Similarly, the scores of cross-quality genuine pairs converge to a point that is higher than the majority of impostor scores. This means the two types of errors discussed in Section~\ref{sec:motivation} could be solved by PFE. This is further confirmed by the IJB-A results in Figure~\ref{fig:ijba_fail_prob}. Figure~\ref{fig:hist_sigma} shows the distribution of estimated uncertainty on LFW, IJB-A and IJB-S. As one can see, the ``variance" of uncertainty increases in the following order: LFW $<$ IJB-A $<$ IJB-S. Comparing with the performance in Section~\ref{sec:exp:msceleb}, we can see that PFE tends to achieve larger performance improvement on datasets with more diverse image quality.\vspace{-1.2em}

\begin{figure}[t]
\setlength\tabcolsep{0px}
\newcommand{\hhh}{33px}
\newcolumntype{Y}{>{\centering\arraybackslash}X}
    \captionsetup{font=footnotesize}
    \footnotesize
    \centering
    \begin{tabularx}{\linewidth}{Ycccccc}
    % \toprule
        $\bx$ & $\bsigma$ & mean & sample1 & sample2 & sample3 & sample4\\
    % \toprule
        \includegraphics[height=\hhh]{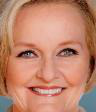} &
        \includegraphics[height=\hhh]{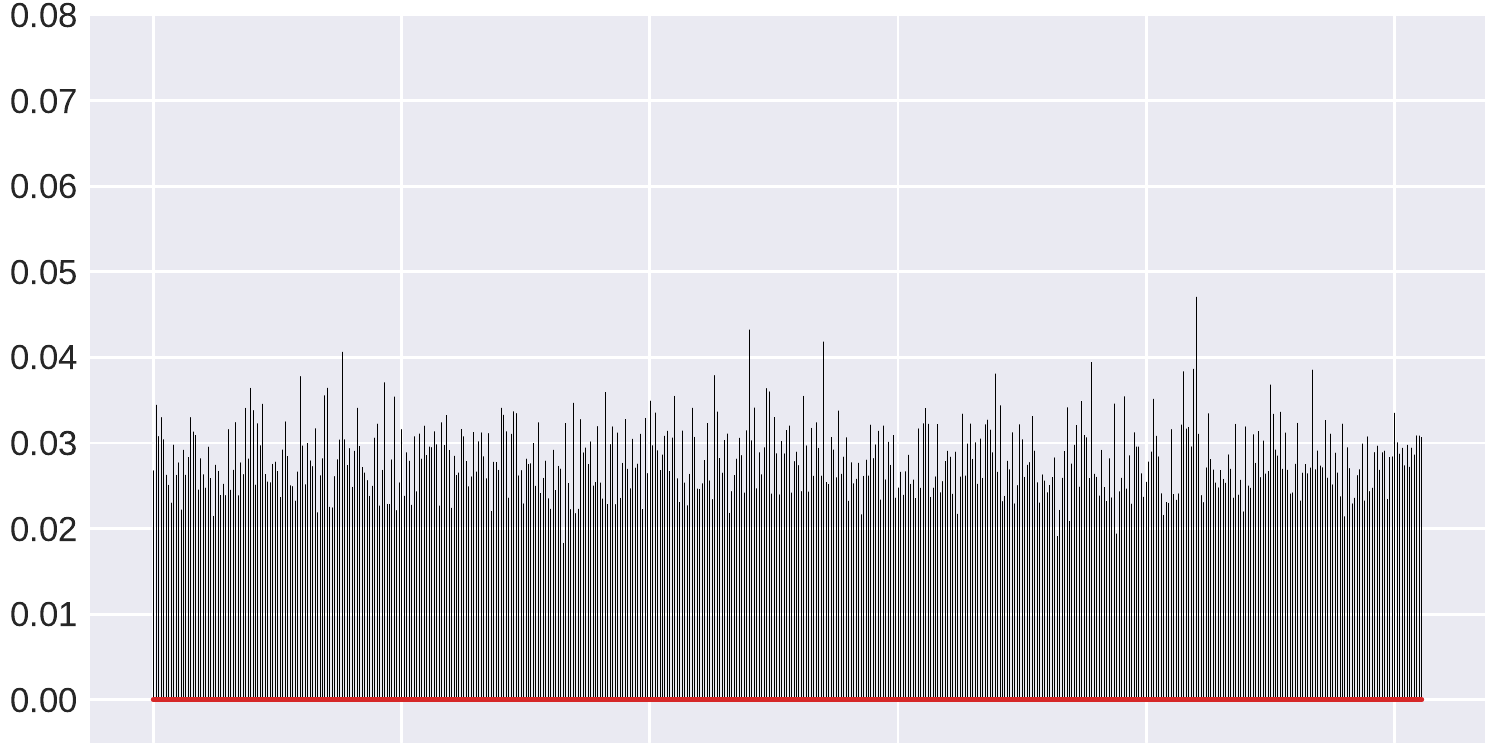}\, &
        \includegraphics[height=\hhh]{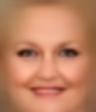} & 
        \includegraphics[height=\hhh]{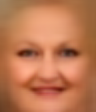} & 
        \includegraphics[height=\hhh]{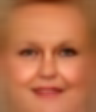} & 
        \includegraphics[height=\hhh]{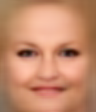} & 
        \includegraphics[height=\hhh]{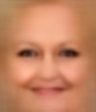} \\[-0.1em]
        \includegraphics[height=\hhh]{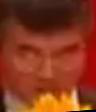} &
        \includegraphics[height=\hhh]{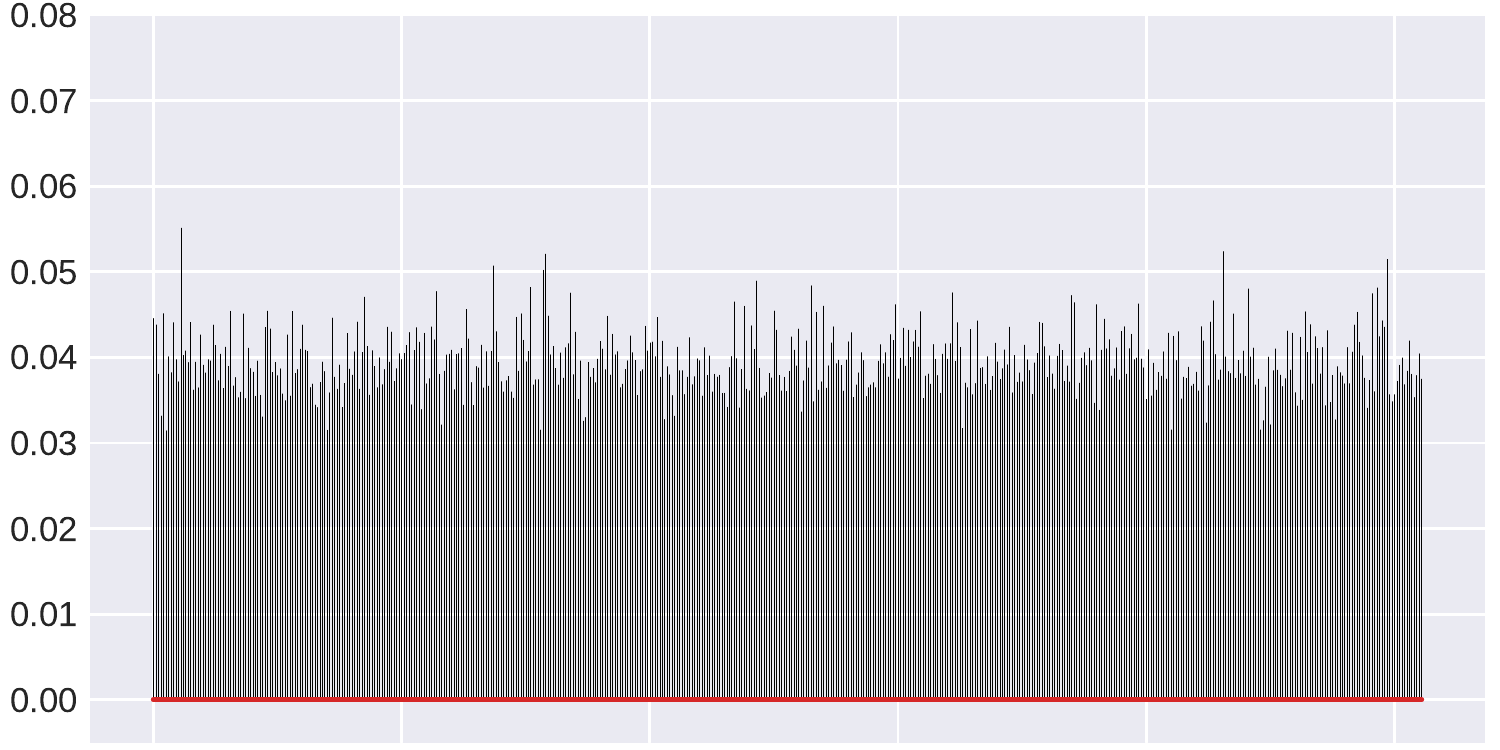}\, &
        \includegraphics[height=\hhh]{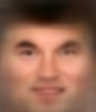} & 
        \includegraphics[height=\hhh]{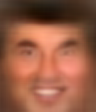} & 
        \includegraphics[height=\hhh]{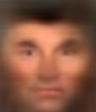} & 
        \includegraphics[height=\hhh]{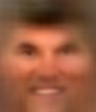} & 
        \includegraphics[height=\hhh]{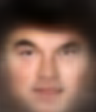} \\[-0.1em]
        \includegraphics[height=\hhh]{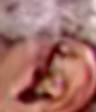} &
        \includegraphics[height=\hhh]{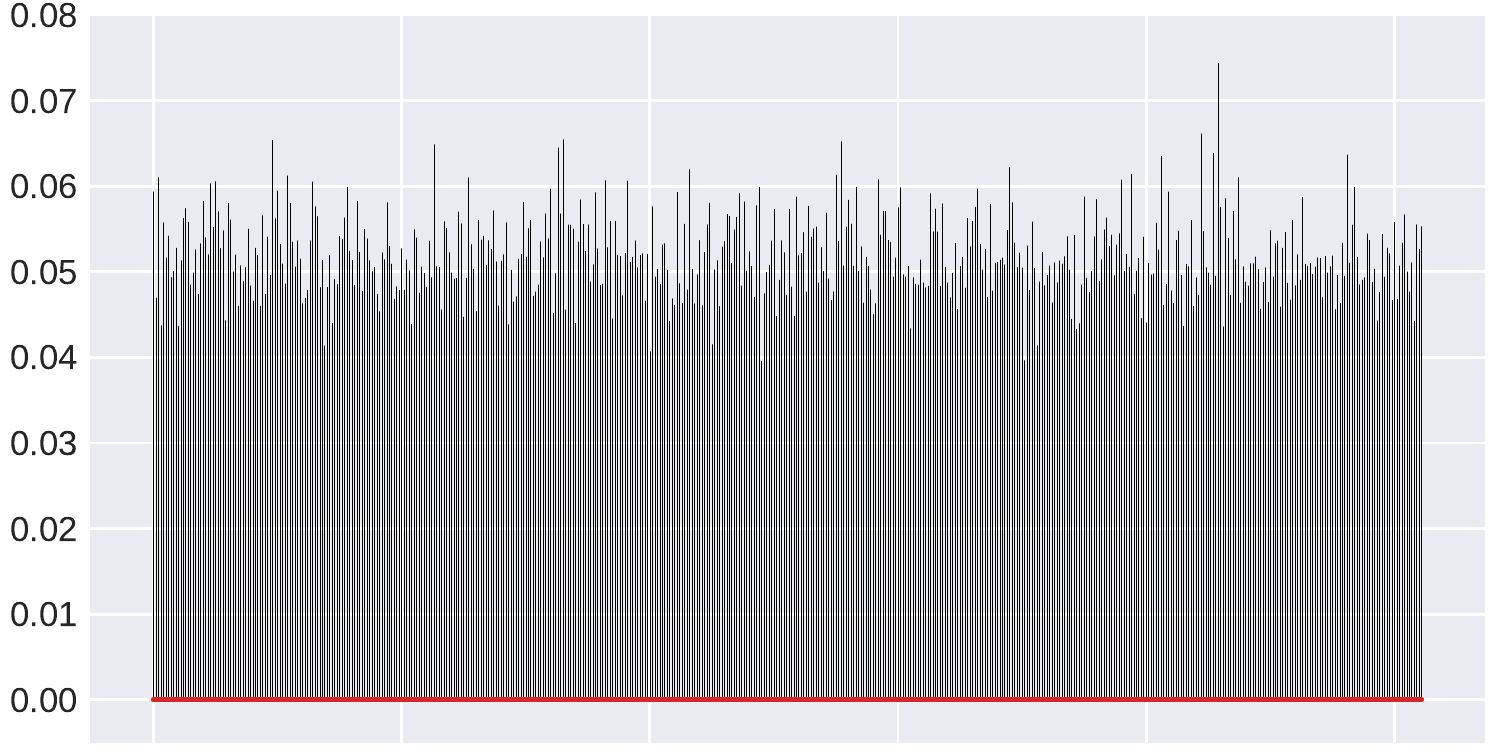}\, &
        \includegraphics[height=\hhh]{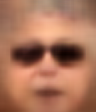} & 
        \includegraphics[height=\hhh]{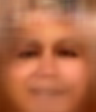} & 
        \includegraphics[height=\hhh]{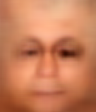} & 
        \includegraphics[height=\hhh]{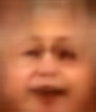} &
        \includegraphics[height=\hhh]{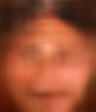} \\[-1.0em]
    % \bottomrule
    \end{tabularx}
    \vspace{-0.2em}\caption{Visualization results on a high-quality, a low-quality and a mis-detected image from IJB-A. For each input, 5 images are reconstructed by a pre-trained decoder using the mean and $4$ randomly sampled $\bz$ vectors from the estimated distribution $p(\bz|\bx)$.}\vspace{-1.1em}
    \label{fig:rec}
\end{figure}

\paragraph{What does DNN see and not see?} To answer this question, we train a decoder network on the original embedding, then apply it to PFE by sampling $\bz$ from the estimated distribution $p(\bz|\bx)$ of given $\bx$. For a high-quality image (Figure~\ref{fig:rec} Row 1), the reconstructed images tend to be very consistent without much variation, implying the model is very certain about the facial features in this images. In contrast, for a lower-quality input (Figure~\ref{fig:rec} Row 2), larger variation can be observed from the reconstructed images. In particular, attributes that can be clearly discerned from the image (\eg thick eye-brow) are still consistent while attributes cannot (\eg eye shape) be discerned have larger variation. As for a mis-detected image (Figure~\ref{fig:rec} Row 3), significant variation can be observed in the reconstructed images: the model does not see any salient feature in the given image.

\section{Risk-controlled Face Recognition}
\begin{figure}[t]
\setlength\tabcolsep{2px}
\newcommand{\hhh}{30px}
\newcolumntype{Y}{>{\centering\arraybackslash}X}
    \captionsetup{font=footnotesize}
    \footnotesize
    \centering
    \begin{tabularx}{\linewidth}{Y>{\centering\arraybackslash}ccc}
       & & LFW &  IJB-A\\
        \multirow{2}{*}{\rotatebox[origin=c]{90}{Ours}} & \raisebox{1.5\height}{H} & 
        \includegraphics[height=\hhh]{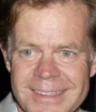}\hfill
        \includegraphics[height=\hhh]{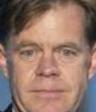}\hfill
        \includegraphics[height=\hhh]{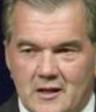}\hfill
        \includegraphics[height=\hhh]{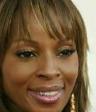} &
        \includegraphics[height=\hhh]{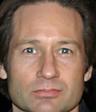}\hfill
        \includegraphics[height=\hhh]{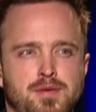}\hfill
        \includegraphics[height=\hhh]{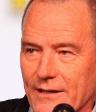}\hfill
        \includegraphics[height=\hhh]{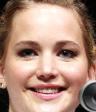} \\[-0.3em]
        & \raisebox{1.5\height}{L} & 
        \includegraphics[height=\hhh]{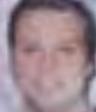}\hfill
        \includegraphics[height=\hhh]{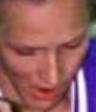}\hfill
        \includegraphics[height=\hhh]{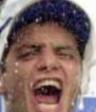}\hfill
        \includegraphics[height=\hhh]{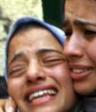} &
        \includegraphics[height=\hhh]{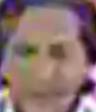}\hfill
        \includegraphics[height=\hhh]{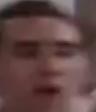}\hfill
        \includegraphics[height=\hhh]{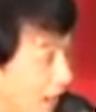}\hfill
        \includegraphics[height=\hhh]{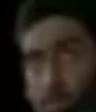} \\
        \multirow{2}{*}{\rotatebox[origin=c]{90}{MTCNN}} & \raisebox{1.5\height}{H} & 
        \includegraphics[height=\hhh]{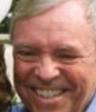}\hfill
        \includegraphics[height=\hhh]{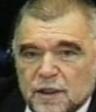}\hfill
        \includegraphics[height=\hhh]{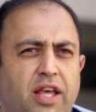}\hfill
        \includegraphics[height=\hhh]{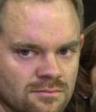} &
        \includegraphics[height=\hhh]{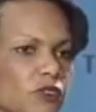}\hfill
        \includegraphics[height=\hhh]{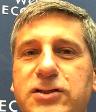}\hfill
        \includegraphics[height=\hhh]{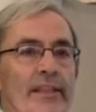}\hfill
        \includegraphics[height=\hhh]{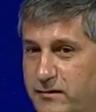} \\[-0.3em]
        & \raisebox{1.5\height}{L} & 
        \includegraphics[height=\hhh]{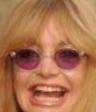}\hfill
        \includegraphics[height=\hhh]{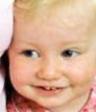}\hfill
        \includegraphics[height=\hhh]{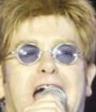}\hfill
        \includegraphics[height=\hhh]{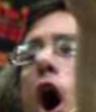} &
        \includegraphics[height=\hhh]{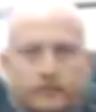}\hfill
        \includegraphics[height=\hhh]{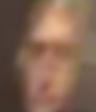}\hfill
        \includegraphics[height=\hhh]{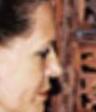}\hfill
        \includegraphics[height=\hhh]{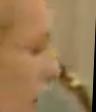} \\[-1.0em]
    \end{tabularx}
    \vspace{-0.3em}\caption{Example images from LFW and IJB-A that are estimated with the highest (H) confidence/quality scores and the lowest (L) scores by our method and MTCNN face detector.}\vspace{-1.7em}
    \label{fig:quality}
\end{figure}

\begin{figure}[t]
    \centering
    \captionsetup{font=footnotesize}
    \begin{minipage}{0.48\linewidth}
    \includegraphics[width=1\linewidth]{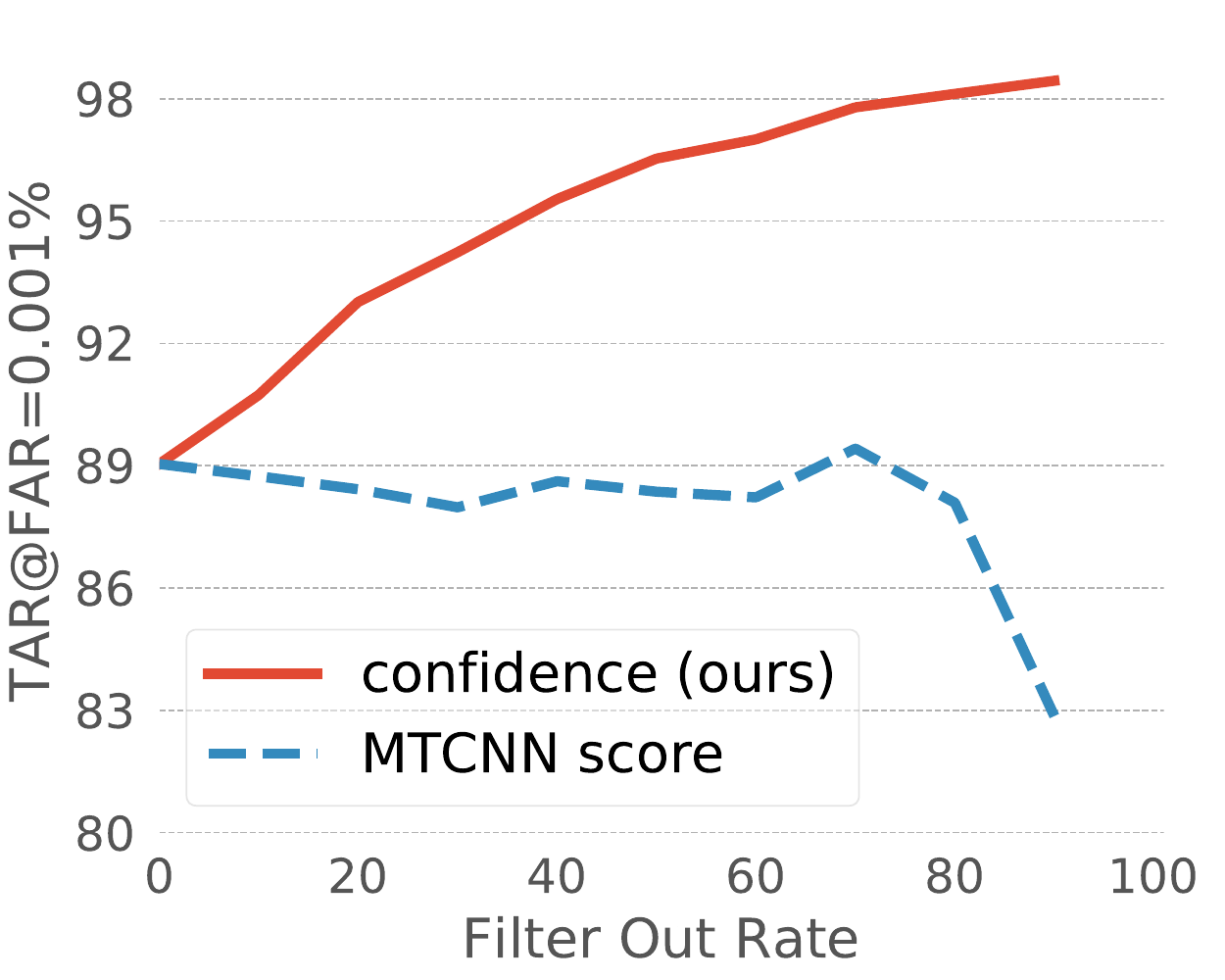}\\[-2.0em]
    \begin{center} \footnotesize(a) LFW \end{center}\vspace{-1.2em}
    \end{minipage}\hfill
    \begin{minipage}{0.48\linewidth}
    \includegraphics[width=1\linewidth]{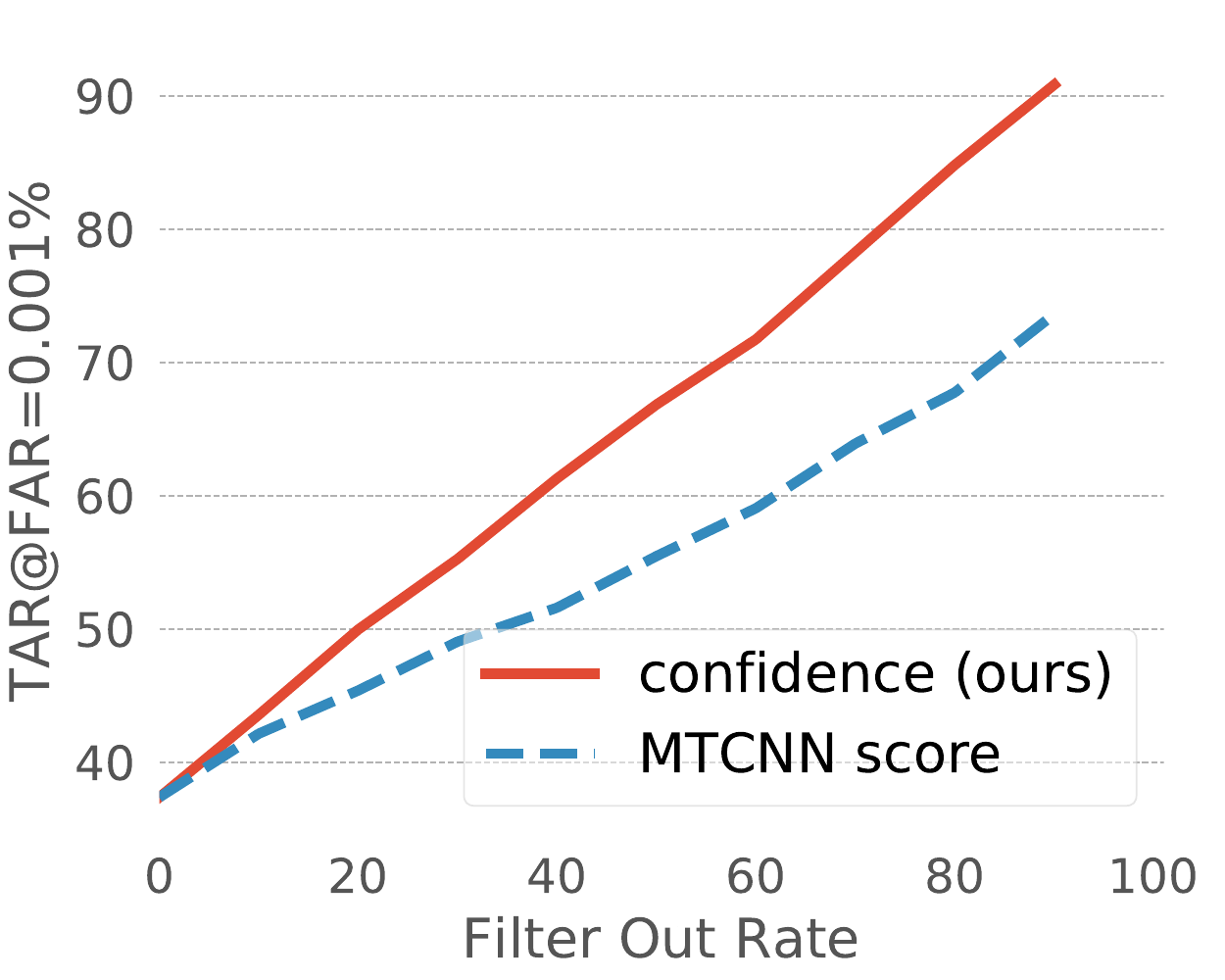}\\[-2.0em]
    \begin{center} \footnotesize(b) IJB-A \end{center}\vspace{-1.2em}
    \end{minipage}\hfill
    \vspace{-0.3em}\caption{ Comparison of verification performance on LFW and IJB-A (not the original protocol) by filtering a proportion of images using different quality criteria.}\vspace{-1.7em}
    \label{fig:FOR}
\end{figure}

In many scenarios, we may expect a higher performance than our system is able to achieve or we may want to make sure the system's performance can be controlled when facing complex application scenarios. Therefore, we would expect the model to reject input images if it is not confident. A common solution for this is to filter the images with a quality assessment tool. We show that PFE provides a natural solution for this task. 
We take all the images from LFW and IJB-A datasets for image-level face verification (We do not follow the original protocols here). The system is allowed to ``filter out'' a proportion of all images to maintain a better performance. We then report the TAR@FAR$=0.001\%$ against the ``Filter Out Rate''. We consider two criteria for filtering: (1) the detection score of MTCNN~\cite{MTCNN} and (2) a confidence value predicted by our uncertainty module. Here the confidence for $i^{\text{th}}$ sample is defined as the inverse of harmonic mean of $\bsigma_i$ across all dimensions. For fairness, both methods use the original deterministic embedding representations and cosine similarity for matching. To avoid saturated results, we use the model trained on CASIA-WebFace with AM-Softmax. The results are shown in Figure~\ref{fig:FOR}. As one can see, the predicted confidence value is a better indicator of the potential recognition accuracy of the input image. This is an expected result since PFE is trained under supervision for the particular model while an external quality estimator is unaware of the kind of features used for matching by the model. Example images with high/low confidence/quality scores are shown in Figure~\ref{fig:quality}.

%------------------------------------------------------------------------
\section{Conclusion}
We have proposed probabilistic face embeddings (PFEs), which represent face images as distributions in the latent space. Probabilistic solutions were derived to compare and aggregate the PFE of face images. Unlike deterministic embeddings, PFEs do not suffer from the feature ambiguity dilemma for unconstrained face recognition. Quantitative and qualitative analysis on different settings showed that PFEs can effectively improve the face recognition performance by converting deterministic embeddings to PFEs. We have also shown that the uncertainty in PFEs is a good indicator for the ``discriminative''quality of face images. In the future work we will explore how to learn PFEs in an end-to-end manner and how to address the data dependency within face templates.

{\small
\bibliographystyle{ieee_fullname}
\bibliography{egbib}

\begin{thebibliography}{10}\itemsep=-1pt

\bibitem{arandjelovic2005face}
Ognjen Arandjelovic, Gregory Shakhnarovich, John Fisher, Roberto Cipolla, and
  Trevor Darrell.
\newblock Face recognition with image sets using manifold density divergence.
\newblock In {\em CVPR}, 2005.

\bibitem{cao2018vggface2}
Qiong Cao, Li Shen, Weidi Xie, Omkar~M Parkhi, and Andrew Zisserman.
\newblock Vggface2: A dataset for recognising faces across pose and age.
\newblock In {\em FG}, 2018.

\bibitem{cevikalp2010face}
Hakan Cevikalp and Bill Triggs.
\newblock Face recognition based on image sets.
\newblock In {\em CVPR}, 2010.

\bibitem{deng2018arcface}
Jiankang Deng, Jia Guo, Niannan Xue, and Stefanos Zafeiriou.
\newblock Arcface: Additive angular margin loss for deep face recognition.
\newblock In {\em CVPR}, 2019.

\bibitem{gal2016dropout}
Yarin Gal and Zoubin Ghahramani.
\newblock Dropout as a bayesian approximation: Representing model uncertainty
  in deep learning.
\newblock In {\em ICML}, 2016.

\bibitem{gong2017capacity}
Sixue Gong, Vishnu~Naresh Boddeti, and Anil~K Jain.
\newblock On the capacity of face representation.
\newblock {\em arXiv:1709.10433}, 2017.

\bibitem{gong2019video}
Sixue Gong, Yichun Shi, and Anil~K Jain.
\newblock Video face recognition: Component-wise feature aggregation network
  (c-fan).
\newblock In {\em ICB}, 2019.

\bibitem{MS-CELEB}
Yandong Guo, Lei Zhang, Yuxiao Hu, Xiaodong He, and Jianfeng Gao.
\newblock Ms-celeb-1m: A dataset and benchmark for large-scale face
  recognition.
\newblock In {\em ECCV}, 2016.

\bibitem{hiremath2007modelling}
PS Hiremath, Ajit Danti, and CJ Prabhakar.
\newblock Modelling uncertainty in representation of facial features for face
  recognition.
\newblock In {\em Face recognition}. 2007.

\bibitem{LFWTech}
Gary~B. Huang, Manu Ramesh, Tamara Berg, and Erik Learned-Miller.
\newblock Labeled faces in the wild: A database for studying face recognition
  in unconstrained environments.
\newblock Technical report, University of Massachusetts, Amherst, 2007.

\bibitem{huang2015log}
Zhiwu Huang, Ruiping Wang, Shiguang Shan, Xianqiu Li, and Xilin Chen.
\newblock Log-euclidean metric learning on symmetric positive definite manifold
  with application to image set classification.
\newblock In {\em ICML}, 2015.

\bibitem{ioffe2015batch}
Sergey Ioffe and Christian Szegedy.
\newblock Batch normalization: Accelerating deep network training by reducing
  internal covariate shift.
\newblock In {\em ICML}, 2015.

\bibitem{IJBS}
Nathan~D. Kalka, Brianna Maze, James~A. Duncan, Kevin~J. O’Connor, Stephen
  Elliott, Kaleb Hebert, Julia Bryan, and Anil~K. Jain.
\newblock {IJB-S : IARPA Janus Surveillance Video Benchmark }.
\newblock In {\em BTAS}, 2018.

\bibitem{kemelmacher2016megaface}
Ira Kemelmacher-Shlizerman, Steven~M Seitz, Daniel Miller, and Evan Brossard.
\newblock The megaface benchmark: 1 million faces for recognition at scale.
\newblock In {\em CVPR}, 2016.

\bibitem{kendall2015bayesian}
Alex Kendall, Vijay Badrinarayanan, and Roberto Cipolla.
\newblock Bayesian segnet: Model uncertainty in deep convolutional
  encoder-decoder architectures for scene understanding.
\newblock In {\em BMVC}, 2015.

\bibitem{kendall2017uncertainties}
Alex Kendall and Yarin Gal.
\newblock What uncertainties do we need in bayesian deep learning for computer
  vision?
\newblock In {\em NIPS}, 2017.

\bibitem{khan2019striking}
Salman Khan, Munawar Hayat, Waqas Zamir, Jianbing Shen, and Ling Shao.
\newblock Striking the right balance with uncertainty.
\newblock {\em arXiv:1901.07590}, 2019.

\bibitem{kingma2013auto}
Diederik~P Kingma and Max Welling.
\newblock Auto-encoding variational bayes.
\newblock In {\em ICLR}, 2013.

\bibitem{IJBA}
Brendan~F Klare, Ben Klein, Emma Taborsky, Austin Blanton, Jordan Cheney,
  Kristen Allen, Patrick Grother, Alan Mah, and Anil~K Jain.
\newblock Pushing the frontiers of unconstrained face detection and
  recognition: {IARPA Janus Benchmark A}.
\newblock In {\em CVPR}, 2015.

\bibitem{li2013probabilistic}
Haoxiang Li, Gang Hua, Zhe Lin, Jonathan Brandt, and Jianchao Yang.
\newblock Probabilistic elastic matching for pose variant face verification.
\newblock In {\em CVPR}, 2013.

\bibitem{liu2017sphereface}
Weiyang Liu, Yandong Wen, Zhiding Yu, Ming Li, Bhiksha Raj, and Le Song.
\newblock Sphereface: Deep hypersphere embedding for face recognition.
\newblock In {\em CVPR}, 2017.

\bibitem{liu2017quality}
Yu Liu, Junjie Yan, and Wanli Ouyang.
\newblock Quality aware network for set to set recognition.
\newblock In {\em CVPR}, 2017.

\bibitem{mackay1992practical}
David~JC MacKay.
\newblock A practical bayesian framework for backpropagation networks.
\newblock {\em Neural Computation}, 1992.

\bibitem{IJBC}
Brianna Maze, Jocelyn Adams, James~A Duncan, Nathan Kalka, Tim Miller, Otto
  Charles, Anil~K Jain, Niggel Tyler, Janet Anderson, Jordan Cheney, and
  Patrick Grother.
\newblock Iarpa janus benchmark-c: Face dataset and protocol.
\newblock In {\em ICB}, 2018.

\bibitem{neal1995bayesian}
Radford~M Neal.
\newblock {\em Bayesian learning for neural networks}.
\newblock PhD thesis, University of Toronto, 1995.

\bibitem{ranjan2017l2}
Rajeev Ranjan, Carlos~D Castillo, and Rama Chellappa.
\newblock L2-constrained softmax loss for discriminative face verification.
\newblock {\em arXiv:1703.09507}, 2017.

\bibitem{sankaranarayanan2016triplet}
Swami Sankaranarayanan, Azadeh Alavi, Carlos~D Castillo, and Rama Chellappa.
\newblock Triplet probabilistic embedding for face verification and clustering.
\newblock In {\em BTAS}, 2016.

\bibitem{schroff2015facenet}
Florian Schroff, Dmitry Kalenichenko, and James Philbin.
\newblock Facenet: A unified embedding for face recognition and clustering.
\newblock In {\em CVPR}, 2015.

\bibitem{CFP}
Soumyadip Sengupta, Jun-Cheng Chen, Carlos Castillo, Vishal~M Patel, Rama
  Chellappa, and David~W Jacobs.
\newblock Frontal to profile face verification in the wild.
\newblock In {\em WACV}, 2016.

\bibitem{shakhnarovich2002face}
Gregory Shakhnarovich, John~W Fisher, and Trevor Darrell.
\newblock Face recognition from long-term observations.
\newblock In {\em ECCV}, 2002.

\bibitem{deepid2plus}
Yi Sun, Xiaogang Wang, and Xiaoou Tang.
\newblock Deeply learned face representations are sparse, selective, and
  robust.
\newblock In {\em CVPR}, 2015.

\bibitem{taigman2014deepface}
Yaniv Taigman, Ming Yang, Marc'Aurelio Ranzato, and Lior Wolf.
\newblock Deepface: Closing the gap to human-level performance in face
  verification.
\newblock In {\em CVPR}, 2014.

\bibitem{tran2017disentangled}
Luan Tran, Xi Yin, and Xiaoming Liu.
\newblock Disentangled representation learning gan for pose-invariant face
  recognition.
\newblock In {\em CVPR}, 2017.

\bibitem{wang2018additive}
Feng Wang, Jian Cheng, Weiyang Liu, and Haijun Liu.
\newblock Additive margin softmax for face verification.
\newblock {\em IEEE Signal Processing Letters}, 2018.

\bibitem{wang2018cosface}
Hao Wang, Yitong Wang, Zheng Zhou, Xing Ji, Dihong Gong, Jingchao Zhou, Zhifeng
  Li, and Wei Liu.
\newblock Cosface: Large margin cosine loss for deep face recognition.
\newblock In {\em CVPR}, 2018.

\bibitem{wen2016discriminative}
Yandong Wen, Kaipeng Zhang, Zhifeng Li, and Yu Qiao.
\newblock A discriminative feature learning approach for deep face recognition.
\newblock In {\em ECCV}, 2016.

\bibitem{MTCNN}
Yandong Wen, Kaipeng Zhang, Zhifeng Li, and Yu Qiao.
\newblock A discriminative feature learning approach for deep face recognition.
\newblock In {\em ECCV}, 2016.

\bibitem{YTF}
Lior Wolf, Tal Hassner, and Itay Maoz.
\newblock Face recognition in unconstrained videos with matched background
  similarity.
\newblock In {\em CVPR}, 2011.

\bibitem{wu2015light}
Xiang Wu, Ran He, Zhenan Sun, and Tieniu Tan.
\newblock A light cnn for deep face representation with noisy labels.
\newblock {\em IEEE Trans. on Information Forensics and Security}, 2015.

\bibitem{xie2018comparator}
Weidi Xie, Li Shen, and Andrew Zisserman.
\newblock Comparator networks.
\newblock In {\em ECCV}, 2018.

\bibitem{xie2018multicolumn}
Weidi Xie and Andrew Zisserman.
\newblock Multicolumn networks for face recognition.
\newblock In {\em BMVC}, 2018.

\bibitem{xu2014data}
Yong Xu, Xiaozhao Fang, Xuelong Li, Jiang Yang, Jane You, Hong Liu, and Shaohua
  Teng.
\newblock Data uncertainty in face recognition.
\newblock {\em IEEE Trans. on Cybernetics}, 2014.

\bibitem{yang2017neural}
Jiaolong Yang, Peiran Ren, Dongqing Zhang, Dong Chen, Fang Wen, Hongdong Li,
  and Gang Hua.
\newblock Neural aggregation network for video face recognition.
\newblock In {\em CVPR}, 2017.

\bibitem{CASIA-WebFace}
Dong Yi, Zhen Lei, Shengcai Liao, and Stan~Z Li.
\newblock Learning face representation from scratch.
\newblock {\em arXiv:1411.7923}, 2014.

\bibitem{yin2018towards}
Bangjie Yin, Luan Tran, Haoxiang Li, Xiaohui Shen, and Xiaoming Liu.
\newblock Towards interpretable face recognition.
\newblock {\em arXiv:1805.00611}, 2018.

\bibitem{yin2018multi}
Xi Yin and Xiaoming Liu.
\newblock Multi-task convolutional neural network for pose-invariant face
  recognition.
\newblock {\em IEEE Trans. on Image Processing}, 2018.

\bibitem{zafar2019face}
Umara Zafar, Mubeen Ghafoor, Tehseen Zia, Ghufran Ahmed, Ahsan Latif,
  Kaleem~Razzaq Malik, and Abdullahi~Mohamud Sharif.
\newblock Face recognition with bayesian convolutional networks for robust
  surveillance systems.
\newblock {\em EURASIP Journal on Image and Video Processing}, 2019.

\end{thebibliography}
}

\appendix
\section{Proofs}
\label{sec:proof}
\subsection{Mutual Likelihood Score}
Here we prove Equation~(3) in the main paper. For simplicity, we do not need to directly solve the integral. Instead, let us consider an alternative vector $\Delta\bz=\bz_i-\bz_j$, where $\bz_i\sim p(\bz|\bx_i)$, $\bz_j\sim p(\bz|\bx_j)$ and $(\bx_i, \bx_j)$ are the pair of images we need to compare. Then, $p(\bz_i=\bz_j)$ ,~\ie Equation~(2) in the main paper, is equivalent to the density value of $p(\Delta \bz=\mathbf{0})$.

The $l^{\text{th}}$ component (dimension) of $\Delta \bz$, $\Delta z^{(l)}$, is the subtraction of two Gaussian variables, which means:
\begin{equation}
    \Delta z^{(l)}\sim \mathcal{N}(\mu_i^{(l)}-\mu_j^{(l)}, \sigma_i^{2(l)} + \sigma_j^{2(l)}).
\end{equation}
Therefore, the mutual likelihood score is given by:
\begin{align}
\begin{split}
     &\; s(\bx_i,\bx_j) \\
    =&\; \log p(\bz_i=\bz_j) \\
    =&\; \log p(\Delta \bz=\mathbf{0}) \\
    =&\; \sum_l^D \log p(\Delta z^{(l)}=0)\\
    =&\; -\frac{1}{2}\sum_{l=1}^{D}(\frac{(\mu^{(l)}_i-\mu^{(l)}_j)^2}{\sigma_i^{2(l)}+\sigma_j^{2(l)}}+\log(\sigma_i^{2(l)}+\sigma_j^{2(l)}))\\
     &\; -\frac{D}{2}\log{2\pi}.
\end{split}
\end{align}
Note that directly solving the integral will lead to the same solution.

\subsection{Property 1}
Let us consider the case that $\sigma_i^{2(l)}$ equals to a constant $c>0$ for any image $\bx_i$ and dimension $l$. Thus the mutual likelihood score between a pair $(\bx_i,\bx_j)$ becomes:
\begin{align}
\begin{split}
     &\; s(\bx_i,\bx_j) \\
    =&\; -\frac{1}{2}\sum_{l=1}^{D}(\frac{(\mu^{(l)}_i-\mu_j^{(l)})^2}{2c}+\log(2c))-\frac{D}{2}\log{2\pi}\\
    =&\; -c_1\norm{\bmu_i-\bmu_j}^2-c_2,\\
\end{split}
\end{align}
where $c_1=\frac{1}{4c}$ and $c_2=\frac{D}{2}\log(4\pi c)$ are both constants.

\subsection{Representation Fusion}
We first prove Equation~(5) in the main paper. Assuming all the observations $\bx_1,\bx_2\dots \bx_{n+1}$ are conditionally independent given the latent code $z$. The posterior distribution is:
\vspace{-0.3em}\begin{align}
\begin{split}
 &\; p(\bz|\bx_1,\bx_2,\dots,\bx_{n+1})\\
=&\; \frac{p(\bx_1,\bx_2,\dots,\bx_{n+1}|\bz)p(\bz)}{p(\bx_1,\bx_2,\dots,\bx_{n+1})}\\
=&\; \frac{p(\bx_1,\bx_2,\dots,\bx_n|\bz)p(\bx_{n+1}|\bz)p(\bz)}{p(\bx_1,\bx_2,\dots,\bx_{n+1})}\\
=&\; \frac{p(\bx_1,\bx_2,\dots,\bx_{n})p(\bx_{n+1})}{p(\bx_1,\bx_2,\dots,\bx_{n+1})}\frac{p(\bx_1,\bx_2,\dots,\bx_n|\bz)p(\bx_{n+1}|\bz)p(\bz)}{p(\bx_1,\bx_2,\dots,\bx_{n})p(\bx_{n+1})}\\
=&\; \alpha\frac{p(\bx_1,\bx_2,\dots,\bx_n,\bz)p(\bx_{n+1},\bz)}{p(\bx_1,\bx_2,\dots,\bx_{n})p(\bx_{n+1})p(\bz)}\\
=&\; \alpha\frac{p(\bz|\bx_{n+1})}{p(\bz)}p(\bz|\bx_{1},\bx_{2},\dots,\bx_{n}),
\end{split}\raisetag{1.2\baselineskip}\label{eq:fusion_p_expand}
\end{align}
% =&\; \frac{p(\bz)}{p(\bx_1,\bx_2,\dots,\bx_{n+1})}\prod_{i=1}^{n+1}{p(\bx_i|\bz)}\\
% =&\; \frac{p(\bz)}{p(\bx_1,\bx_2,\dots,\bx_{n+1})}\prod_{i=1}^{n+1}{\frac{p(\bz|\bx_i)p(\bx_i)}{p(\bz)}}\\
% =&\; \frac{\prod_{i=1}^{n+1}{p(\bx_i)}}{p(\bx_1,\bx_2,\dots,\bx_{n+1})}\frac{\prod_{i=1}^{n+1}{p(\bz|\bx_i)}}{p(\bz)^n}\\
% =&\; \alpha\frac{\prod_{i=1}^{n+1}{p(\bz|\bx_i)}}{p(\bz)^n}\\
% =&\; \frac{p(\bx_{n+1}|\bz)}{p(\bx_{n+1})}\frac{p(\bx_1,\bx_2,\dots,\bx_n|\bz)p(\bz)}{p(\bx_1,\bx_2,\dots,\bx_{n})}\\
% =&\; \frac{p(\bz|\bx_{n+1})}{p(\bz)}p(\bz|\bx_{1},\bx_{2},\dots,\bx_{n}).
where $\alpha$ is a normalization constant. In this case, $\alpha=\frac{p(\bx_1,\bx_2,\dots,\bx_{n})p(\bx_{n+1})}{p(\bx_1,\bx_2,\dots,\bx_{n+1})}$. 

Without loss of generality, let us consider a one-dimensional case for the followings. The solution can be easily extended to a multivariate case since all feature dimensions are assumed to be independent. It can be shown that the posterior distribution in Equation~(\ref{eq:fusion_p_expand}) is a Gaussian distribution through induction. Let us assume $p(z|\bx_1,\bx_2,\dots,\bx_n)$ is a Gaussian distribution with $\hat{\mu}_n$ and $\hat{\sigma}^2_n$ as mean and variance, respectively. Note that the initial case $p(z|\bx_1)$ is guaranteed to be a Gaussian distribution.
Let $\mu_0$ and $\sigma^2_0$ denote the parameters of the noninformative prior of $z$. Then, if we take $\log$ on both side of Equation~(\ref{eq:fusion_p_expand}), we have:
\vspace{-0.3em}\begin{align}
\begin{split}
 &\; \log p(z|\bx_1,\bx_2,\dots,\bx_{n+1})\\
=&\; \log{p(z|\bx_{n+1})}+\log{p(z|\bx_{1},\bx_{2},\dots,\bx_{n})}-\log{p(z)}+\text{const}\\
=&\; -\frac{(z-\mu_{n+1})^2}{2\sigma^2_{n+1}}-\frac{(z-\hmu_n)^2}{2\hsigma^2_n} +\frac{(z-\mu_0)^2}{2\sigma^2_{0}} + \text{const}\\
=&\; -\frac{(z-\hmu_{n+1})^2}{2\hsigma^2_{n+1}} + \text{const}.
\end{split}\label{eq:fusion_p_expand_log}
\end{align}
where ``const'' refers to the terms irrelevant to $z$ and
\vspace{-0.3em}\begin{gather}
\label{eq:fusion_conjugate_prior}
\hmu_{n+1}=\hsigma^2_{n+1}(\frac{\mu_{n+1}}{\sigma^2_{n+1}}+\frac{\hmu_n}{\hsigma^2_n}-\frac{\mu_0}{\sigma^2_0}), \\
\frac{1}{\hsigma^2_{n+1}}=\frac{1}{\sigma^2_{n+1}}+\frac{1}{\hsigma^2_{n}}-\frac{1}{\sigma^2_{0}}.
\end{gather}
Considering $\sigma_0 \rightarrow \infty$, we have
\vspace{-0.3em}\begin{gather}
\label{eq:fusion_conjugate}
\hmu_{n+1}=\frac{\hsigma^2_n\mu_{n+1}+\sigma^2_{n+1}\hmu_n}{\sigma^2_{n+1}+\hsigma^2_n}, \\
\hsigma^2_{n+1}=\frac{\sigma^2_{n+1}\hsigma^2_n}{\sigma^2_{n+1}+\hsigma^2_n}.
\end{gather}
The result means the posterior distribution is a new Gaussian distribution with a smaller variance. Further, we can directly give the solution of fusing $n$ samples:
\vspace{-0.3em}\begin{align}
\begin{split}
 &\; \log p(z|\bx_1,\bx_2,\dots,\bx_{n})\\
=&\; \log [\alpha p(z|\bx_1)\prod_{i=2}^{n}{\frac{p(z|\bx_i)}{p(z)}}]\\
=&\; (n-1){\log{p(z)}}-\sum_{i=1}^{n}{\log{p(z|\bx_{i})}} + \text{const}\\
=&\; (n-1)\frac{(z-\mu_0)^2}{2\sigma^2_{0}}-\sum_{i=1}^{n}{\frac{(z-\mu_{i})^2}{2\sigma^2_{i}}} + \text{const}\\
% =&\; \sum_{i=1}^{n}{\frac{z^2}{2\sigma^2_{i}}-\sum_{i=1}^{n}{\frac{z\mu_i}{\sigma^2_{i}}}} + \text{const}\\
=&\; -\frac{(z-\hmu_{n})^2}{2\hsigma^2_{n}} + \text{const}.
\end{split}\label{eq:fusion_all_log}
\end{align}
where $\alpha=\frac{\prod_{i=1}^{n}{p(\bx_i)}}{p(\bx_1,\bx_2,\dots,\bx_n)}$ and
\vspace{-0.3em}\begin{gather}
\label{eq:fusion_all_conjugate_prior}
\hmu_{n}=\sum_{i=1}^{n}{\frac{\hsigma^2_n}{\sigma^2_i}\mu_i}-(n-1){\frac{\hsigma^2_n}{\sigma^2_0}\mu_0}, \\
\frac{1}{\hsigma^2_{n}}=\sum_{i=1}^{n}{\frac{1}{\sigma^2_i}}-(n-1)\frac{1}{\sigma^2_0}.
\end{gather}
Considering $\sigma_0 \rightarrow \infty$, we have
\begin{align}
    \hmu_n=\sum_{i=1}^{n}{\frac{\hsigma^2_n}{\sigma^2_i}\mu_i}, \\
    \hsigma^2_n=\frac{1}{\sum_{i=1}^{n}{\frac{1}{\sigma^2_i}}}. 
\end{align}

\section{Implementation Details}
\label{sec:detail}
All the models in the paper are implemented using Tensorflow r1.9. Two and Four GeForce GTX 1080 Ti GPUs are used for training base models on CASIA-Webface~\cite{CASIA-WebFace} and MS-Celeb-1M~\cite{MS-CELEB}, respectively. The uncertainty modules are trained using one GPU.

\subsection{Data Preprocessing}
All the face images are first passed through MTCNN face detector~\cite{MTCNN} to detect 5 facial landmarks (two eyes, nose and two mouth corners). Then, similarity transformation is used to normalize the face images based on the five landmarks. After transformation, the images are resized to $112\times96$. Before passing into networks, each pixel in the RGB image is normalized by subtracting $127.5$ and dividing by $128$. 

\subsection{Base Models}
The base models for CASIA-Webface~\cite{CASIA-WebFace} are trained for $28,000$ steps using a SGD optimizer with a momentum of $0.9$. The learning rate starts at $0.1$, and is decreased to $0.01$ and $0.001$ after $16,000$ and $24,000$ steps, respectively. For the base model trained on Ms-Celeb-1M~\cite{MS-CELEB}, we train the network for $140,000$ steps using the same optimizer settings. The learning rate starts at $0.1$, and is decreased to $0.01$ and $0.001$ after $80,000$ and $120,000$ steps, respectively. The batch size, feature dimension and weight decay are set to $256$, $512$ and $0.0005$, respectively, for both cases.

\subsection{Uncertainty Module}
\paragraph{Architecture} The uncertainty module for all models are two-layer perceptrons with the same architecture: \texttt{FC-BN-ReLU-FC-BN-exp}, where \texttt{FC} refers to fully connected layers, \texttt{BN} refers to batch normalization layers~\cite{ioffe2015batch} and \texttt{exp} function ensures the outputs $\sigma^2$ are all positive values~\cite{kendall2017uncertainties}. The first \texttt{FC} shares the same input with the bottleneck layer,~\ie the output feature map of the last convolutional layer. The output of both \texttt{FC} layers are $D$-dimensional vectors, where $D$ is the dimensionality of the latent space. In addition, we constrain the last \texttt{BN} layer to share the same $\gamma$ and $\beta$ across all dimensions, which we found to help stabilizing the training. 

\paragraph{Training} For the models trained on CASIA-WebFace~\cite{CASIA-WebFace}, we train the uncertainty module for $3,000$ steps using a SGD optimizer with a momentum of $0.9$. The learning rate starts at $0.001$, and is decreased to $0.0001$ after $2,000$ steps. For the model trained on MS-Celeb-1M\cite{MS-CELEB}, we train the uncertainty module for $12,000$ steps. The learning rate starts at $0.001$, and is decreased to $0.0001$ after $8,000$ steps. The batch size for both cases are $256$. For each mini-batch, we randomly select $4$ images from $64$ different subjects to compose the positive pairs ($384$ pairs in all). The weight decay is set to $0.0005$ in all cases. A Subset of the training data was separated as the validation set for choosing these hyper-parameters during development phase. 

\paragraph{Inference Speed} Feature extraction (passing through the whole network) using one GPU takes $1.5$ms per image. Note that given the small size of the uncertainty module, it has little impact on the feature extraction time. Matching images using cosine similarity and mutual likelihood score takes $4$ns and  $15$ns
, respectively. Both are neglectable in comparison with feature extraction time.

\begin{table}[t]
\captionsetup{font=footnotesize}
\footnotesize
\setlength{\tabcolsep}{3pt}
\begin{center}
\begin{tabularx}{\linewidth}{Xccccc}
\toprule
Base Model                      & Representation & LFW & YTF & CFP-FP & IJB-A \\
\midrule
                                                        & Original  & $97.70$ & $92.56$ & $91.13$ & $63.93$\\
\multirow{-2}{*}{\shortstack[l]{Softmax + \\Center Loss}~\cite{wen2016discriminative}}  
                                                        & PFE       & $\bold{97.89}$ & $\bold{93.10}$ & $\bold{91.36}$ & $\bold{64.33}$ \\\rowcolor{Gray}
%%%%%%%%%%%%%%%%%%%%%%%%%%%%%%%%%%%%%%%%%%%%%%%%%%%%%%%%%%%%%%%%%%%%%%%%%%%%%%%%%%%%%%%%%%%%%%%%%%%%%%%%%%%%%%
                                                        & Original  & $96.98$ & $90.72$ & $\bold{85.69}$ & $\bold{54.47}$ \\\rowcolor{Gray}
\multirow{-2}{*}{Triplet~\cite{schroff2015facenet}}     & PFE       & $\bold{97.10}$ & $\bold{91.22}$ & $85.10$ & $51.35$ \\
%%%%%%%%%%%%%%%%%%%%%%%%%%%%%%%%%%%%%%%%%%%%%%%%%%%%%%%%%%%%%%%%%%%%%%%%%%%%%%%%%%%%%%%%%%%%%%%%%%%%%%%%%%%%%%
                                                        & Original  & $97.12$ & $\bold{92.38}$ & $89.31$ & $54.48$ \\
\multirow{-2}{*}{A-Softmax~\cite{liu2017sphereface}}      & PFE       & $\bold{97.92}$ & $91.78$ & $\bold{89.96}$ & $\bold{58.09}$ \\\rowcolor{Gray}
%%%%%%%%%%%%%%%%%%%%%%%%%%%%%%%%%%%%%%%%%%%%%%%%%%%%%%%%%%%%%%%%%%%%%%%%%%%%%%%%%%%%%%%%%%%%%%%%%%%%%%%%%%%%%%
                                                        & Original  & $98.32$ & $93.50$ & $90.24$ & $71.28$ \\\rowcolor{Gray}
\multirow{-2}{*}{AM-Softmax~\cite{wang2018additive}}    & PFE       & $\bold{98.63}$ & $\bold{94.00}$ & $\bold{90.50}$ & $\bold{75.92}$ \\

\bottomrule
\end{tabularx}
\vspace{-0.9em}\caption{Results of CASIA-Net models trained on CASIA-WebFace. ``Original'' refers to the deterministic embeddings. The better performance among each base model are shown in bold numbers. ``PFE'' uses mutual likelihood score for matching. IJB-A results are verification rates at FAR=$0.1\%$.}\vspace{-2.5em}
\label{tab:loss_function_casianet}
\end{center}
\end{table}

\begin{table}[t]
\captionsetup{font=footnotesize}
\footnotesize
\setlength{\tabcolsep}{3pt}
\begin{center}
\begin{tabularx}{\linewidth}{Xccccc}
\toprule
Base Model                      & Representation & LFW & YTF & CFP-FP & IJB-A \\
\midrule
                                                        & Original  & $97.77$ & $92.34$ & $90.96$ & $60.42$\\
\multirow{-2}{*}{\shortstack[l]{Softmax + \\Center Loss}~\cite{wen2016discriminative}}  
                                                        & PFE       & $\bold{98.28}$ & $\bold{93.24}$ & $\bold{92.29}$ & $\bold{62.41}$ \\\rowcolor{Gray}
%%%%%%%%%%%%%%%%%%%%%%%%%%%%%%%%%%%%%%%%%%%%%%%%%%%%%%%%%%%%%%%%%%%%%%%%%%%%%%%%%%%%%%%%%%%%%%%%%%%%%%%%%%%%%%
                                                        & Original  & $97.48$ & $92.46$ & $90.01$ & $52.34$ \\\rowcolor{Gray}
\multirow{-2}{*}{Triplet~\cite{schroff2015facenet}}     & PFE       & $\bold{98.15}$ & $\bold{93.62}$ & $\bold{90.54}$ & $\bold{56.81}$ \\
%%%%%%%%%%%%%%%%%%%%%%%%%%%%%%%%%%%%%%%%%%%%%%%%%%%%%%%%%%%%%%%%%%%%%%%%%%%%%%%%%%%%%%%%%%%%%%%%%%%%%%%%%%%%%%
                                                        & Original  & $98.07$ & $92.72$ & $89.34$ & $63.21$ \\
\multirow{-2}{*}{A-Softmax~\cite{liu2017sphereface}}      & PFE       & $\bold{98.47}$ & $\bold{93.44}$ & $\bold{90.54}$ & $\bold{71.96}$ \\\rowcolor{Gray}
%%%%%%%%%%%%%%%%%%%%%%%%%%%%%%%%%%%%%%%%%%%%%%%%%%%%%%%%%%%%%%%%%%%%%%%%%%%%%%%%%%%%%%%%%%%%%%%%%%%%%%%%%%%%%%
                                                        & Original  & $98.68$ & $93.78$ & $90.59$ & $76.50$ \\\rowcolor{Gray}
\multirow{-2}{*}{AM-Softmax~\cite{wang2018additive}}    & PFE       & $\bold{98.95}$ & $\bold{94.34}$ & $\bold{91.26}$ & $\bold{80.00}$ \\
\bottomrule
\end{tabularx}
\vspace{-0.9em}\caption{Results of Light-CNN models trained on CASIA-WebFace. ``Original'' refers to the deterministic embeddings. The better performance among each base model are shown in bold numbers. ``PFE'' uses mutual likelihood score for matching. IJB-A results are verification rates at FAR=$0.1\%$.}\vspace{-2.5em}
\label{tab:loss_functions_lightcnn}
\end{center}
\end{table}

\section{Results on Different Architectures}
Throughout the main paper, we conducted the experiments using a 64-layer CNN network~\cite{liu2017sphereface}. Here, we evaluate the proposed method on two different network architectures for face recognition: CASIA-Net~\cite{CASIA-WebFace} and 29-layer Light-CNN~\cite{wu2015light}. Notice that both networks require different image shapes from our  preprocessed ones. Thus we pad our images with zero values and resize them into the target size. Since the main purpose of the experiment is to evaluate the efficacy of the uncertainty module rather than comparing with the original results of these networks, the difference in the preprocessing should not affect a fair comparison. Besides, the original CASIA-Net does not converge for A-Softmax and AM-Softmax, so we add an bottleneck layer to output the embedding representation after the average pooling layer. Then we conduct the experiments by comparing probabilistic embeddings with base deterministic embeddings, similar to Section 5.1 in the main paper. The results are shown in Table~\ref{tab:loss_function_casianet} and Table~\ref{tab:loss_functions_lightcnn}. Without tuning the architecture of the uncertainty module nor the hyper-parameters, PFE still improve the performance in most cases.

\end{document}